\documentclass{llncs}
\usepackage{graphicx}
\usepackage{comment}
\usepackage{amsmath,amssymb} 
\usepackage{color}

\newcommand{\etal}{\textit{et al}.~}

\newcommand{\ieno}{\textit{i}.\textit{e}.}

\newcommand{\egno}{\textit{e}.\textit{g}.} 

\usepackage{enumitem}
\usepackage{caption}
\usepackage{comment}
\usepackage{xcolor}
\usepackage{amsfonts}

\usepackage{algorithmic}            
\usepackage{algorithm}
\usepackage{amsmath}
\usepackage{booktabs}       
\usepackage{multirow}
\usepackage{amsmath}
\usepackage{statex}
\makeatletter
\def\wideubar{\underaccent{{\cc@style\underline{\mskip10mu}}}}
\def\Wideubar{\underaccent{{\cc@style\underline{\mskip8mu}}}}
\makeatother
\usepackage{colortbl}
\usepackage{graphicx}
\usepackage{float}
\usepackage{graphicx}
\usepackage[lofdepth,lotdepth]{subfig}
\newcommand{\tablestyle}[2]{\setlength{\tabcolsep}{#1}\renewcommand{\arraystretch}{#2}\centering\scriptsize}

\makeatletter
\def\widebar{\accentset{{\cc@style\underline{\mskip10mu}}}}
\def\Widebar{\accentset{{\cc@style\underline{\mskip8mu}}}}
\makeatother

\begin{document}
\pagestyle{headings}

\title{Global Distance-distributions Separation for Unsupervised Person Re-identification} 


\titlerunning{Abbreviated paper title}
\authorrunning{F. Author et al.}

\author{{Xin Jin{$^{1,2}$}\thanks{This work was done when Xin Jin was an intern at MSRA.}} \qquad Cuiling Lan{$^{2}$}\thanks{Corresponding Author.} \qquad   Wenjun Zeng{$^{2}$} \qquad  Zhibo Chen{$^{1**}$} \\
}

\institute{University of Science and Technology of China \and
Microsoft Research Asia, Beijing, China\\
\email{jinxustc@mail.ustc.edu.cn\quad \{culan,wezeng\}@microsoft.com\quad chenzhibo@ustc.edu.cn}
}


\maketitle

\begin{abstract}
Supervised person re-identification (ReID) often has poor scalability and usability in real-world deployments due to domain gaps and the lack of annotations for the target domain data. Unsupervised person ReID through domain adaptation is attractive yet challenging. Existing unsupervised ReID approaches often fail in correctly identifying the positive samples and negative samples through the distance-based matching/ranking. The two distributions of distances for positive sample pairs (Pos-distr) and negative sample pairs (Neg-distr) are often not well separated, having large overlap. To address this problem, we introduce a global distance-distributions separation (GDS) constraint over the two distributions to encourage the clear separation of positive and negative samples from a global view. We model the two global distance distributions as Gaussian distributions and push apart the two distributions while encouraging their sharpness in the unsupervised training process. Particularly, to model the distributions from a global view and facilitate the timely updating of the distributions and the GDS related losses, we leverage a momentum update mechanism for building and maintaining the distribution parameters (mean and variance) and calculate the loss on the fly during the training. Distribution-based hard mining is proposed to further promote the separation of the two distributions. We validate the effectiveness of the GDS constraint in unsupervised ReID networks. Extensive experiments on multiple ReID benchmark datasets show our method leads to significant improvement over the baselines and achieves the state-of-the-art performance.

\end{abstract}

\section{Introduction}

\begin{figure*}[t]
  \centerline{\includegraphics[width=0.85\linewidth]{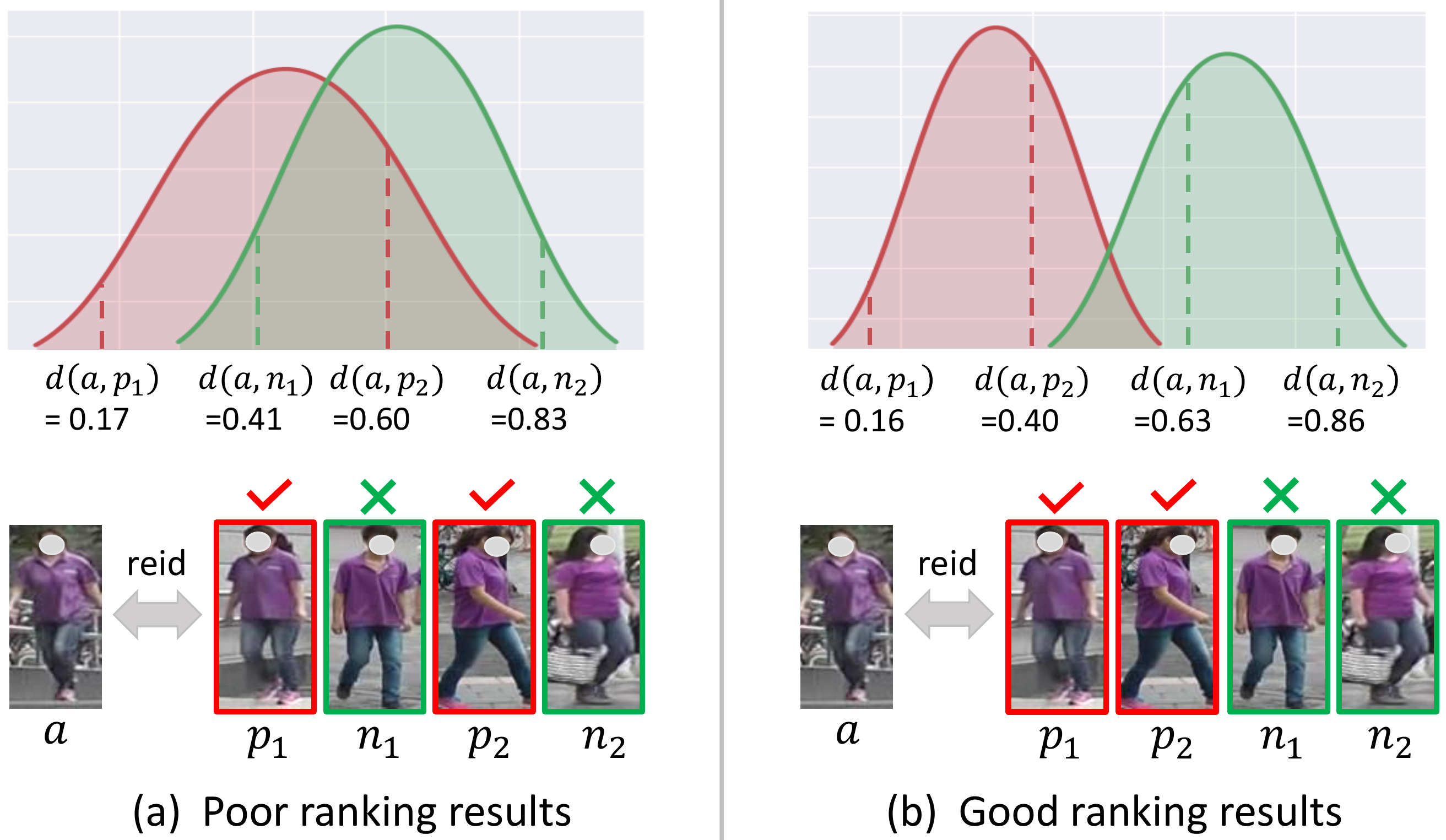}}
  \vspace{-2mm}
  \caption{Illustration of our motivation. The red curves denote the distribution of the distances of positive sample pairs (shorted as Pos-distr) while the green curves denote that of negative sample pairs (shorted as Neg-distr). (a) The two distributions are often not well separated for the current unsupervised ReID models, resulting in poor retrieval/ranking results. (b) Once the two distance distributions are well separated, the positive and negative samples can be well identified based on distance and superior ranking results can be obtained.}
\label{fig:motivation}
\vspace{-6mm}
\end{figure*}

\begin{figure*}[t]
  \centerline{\includegraphics[width=0.95\linewidth]{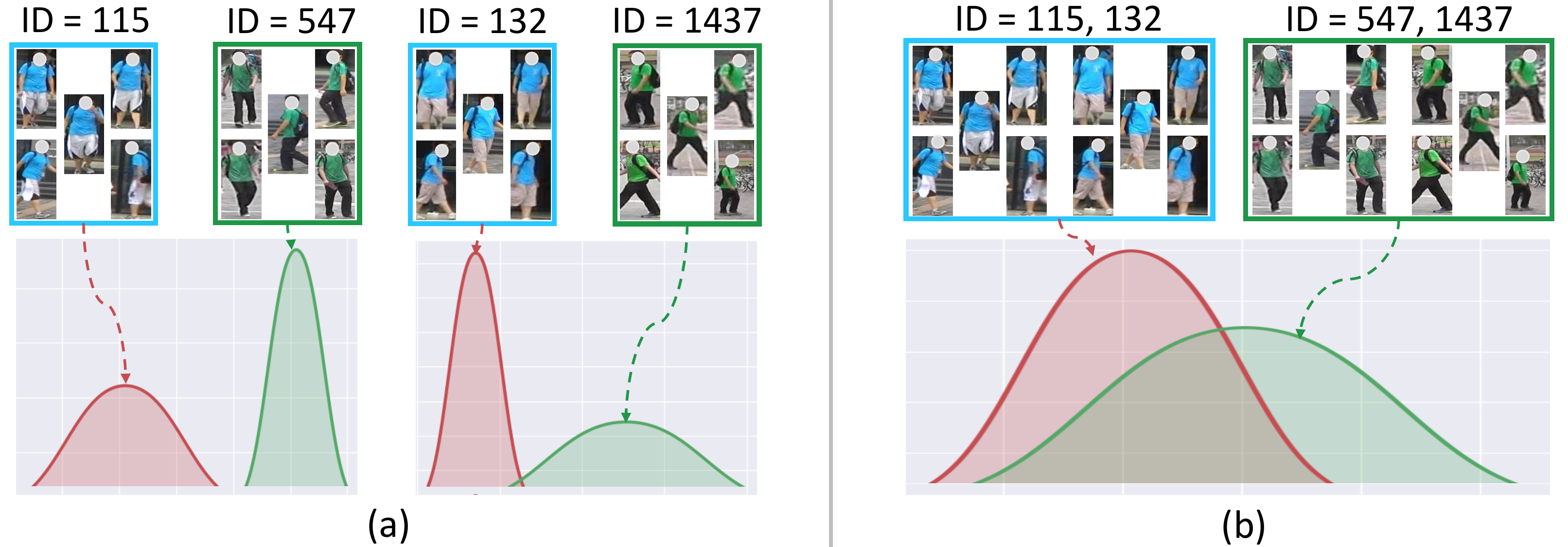}}
  \vspace{-2mm}
  \caption{Local separability is not enough. (a) Within each batch (or subset), the Pos-distr and Neg-distr are well separated. (b) The distributions are inseparable when mixing the two batches together.}
\label{fig:motivation2}
\vspace{-6mm}
\end{figure*}

Person re-identification aims to identify the same person across images. In recent years, significant progress has been made on fully supervised person re-identification (ReID) \cite{su2017pose,zhao2017spindle,ge2018fd,qian2018pose,sun2018beyond,zhang2019DSA,jia2020similarity,jin2020semantics,jin2020uncertainty}, where groundtruth labels are accessible for training. However, they often have poor scalability and usability in real-world deployments. First, they typically perform well on the trained dataset but suffer from signiﬁcant performance degradation when testing on a previously unseen dataset due to the domain gaps. There are usually large style discrepancies across domains/datasets, due to the discrepancy of imaging devices and environments (\egno, lighting conditions, background, viewing angles) \cite{liu2019adaptive}. Second, it is costly to annotate images for each newly deployed environment. One popular solution is unsupervised domain adaptation, which transfers the knowledge from the labeled source domain to the unlabeled target domain. 



Many efforts have been made to develop unsupervised domain adaptation for person re-identification (UDA-ReID) \cite{li2018unsupervised,wang2018transferable,tang2019unsupervised,liu2019adaptive,qi2019novel,fan2018unsupervised,yu2019unsupervised,yang2019patch,zhai2020ad,zhai2020multiple}. Pseudo label based approaches usually pre-train the network with source domain data and predict pseudo labels for the unlabeled target images, \egno, by clustering, followed by fine-tuning with the pseudo labels \cite{yu2017cross,yu2018unsupervised,fan2018unsupervised,song2020unsupervised,zhang2019self,Fu2019SelfsimilarityGA,yang2019asymmetric}. Transfer-based approaches often transfer the labeled source images to have the style of the target domain \cite{wei2018person,deng2018image,liu2019adaptive}. These approaches suffer from either noisy labels or noisy images due to incorrect pseudo labels or unrealisticness of the transferred images \cite{yang2019asymmetric}. Such distractions lead to serious inseparability of positive and negative samples for ReID. 

Person ReID inference/test can be considered as a retrieval task, which aims to identify the images of the same person from a gallery image set by comparing feature distances between the query image and the gallery images \cite{zheng2016person,ye2020deep}. Fig. \ref{fig:motivation}(a) illustrates the two distributions of distances of positive sample pairs (red curve) and negative sample pairs (green curve). The inseparability of the distance distributions leads to poor retrieval/ranking results. Given an anchor image $a$, as shown in Fig. \ref{fig:motivation}(a), its distance ($d(a,n_1)=0.41$) to the negative sample $n_1$ is even smaller than its distance ($d(a,p_2)=0.60$) to the positive sample $p_2$. In comparison, when the two distance distributions are better separated as illustrated in Fig. \ref{fig:motivation}(b), the positive samples can be correctly ranked.




Such poor separability is often observed for unsupervised ReID but is unfortunately under-explored. There is a mismatch between the optimization objective design and the ReID purpose. For ReID inference, it is a \emph{global} ranking/retrieval problem (\egno, based on the feature distances to the query image). The separability of the positive and negative global distance distributions is important for this \emph{global} ranking problem. However, most of the current ReID optimizations/losses are designed from a local perspective. Triplet-based losses \cite{hermans2017defense} optimize the embedding space to encourage the distance of the negative pair to be larger than the positive pair. Such constraint is enforced in three instances, \ieno, (an anchor sample, a positive sample, and a negative sample). However, as shown in the examples in Fig. \ref{fig:motivation}(a), both the triplets of ($a$, $p_1$, $n_1$), and ($a$, $p_2$, $n_2$) meet the marginal constraints but the identification of positive/negative samples fails when ranking them all together. In practice, it is impossible to optimize over all possible triplets of a dataset within an acceptable duration. We argue that dataset-wise (\ieno, global) constraint is imperative to address this problem. Ustinova \etal go a step further and propose a Histogram loss for deep embedding to encourage the separation of distributions within each batch \cite{ustinova2016learning}. Unfortunately, as illustrated in Fig.~\ref{fig:motivation2}, being still a local solution, the separation is easily broken across batches and still cannot guarantee the superiority of the global retrieval performance. Kumar \etal use a global loss to optimize the separation of dataset-wise distributions for learning local image descriptors~\cite{kumar2016learning}. However, their design is less efficient which prohibits the timely update of distributions and leads to inaccurate loss calculation.

In this paper, we propose to optimize unsupervised person ReID from a global distance-distribution perspective by introducing a global distance-distributions separation (GDS) constraint for effective ReID feature learning. Different from the local constraints (\egno, triplet loss) for learning embedding features, we model the global (dataset-wise) distributions of the distances for positive pairs (Pos-distr) and  negative pairs (Neg-distr) and promote their clear separability. Particularly, we model the two global distance distributions as Gaussian distributions with \emph{updatable mean and variance}. To model the distributions from a global view and ensure timely updates of the distributions and GDS-related losses, we leverage the momentum update mechanism for building and maintaining the distribution variables and calculate the GDS loss on the fly during the training. The GDS loss helps push away the two distributions while encouraging the sharpness of individual distribution. Moreover, to better separate the Pos-distr and Neg-distr, we introduce a distribution-based hard mining by pushing the right-tail of the Pos-distr and the left-tail of the Neg-distr to have a soft margin. We summarize our main contributions as follows:

\begin{itemize}[leftmargin=*,noitemsep,nolistsep]

\item We are the first to propose to optimize unsupervised person ReID from a global distance-distribution perspective by encouraging the global separation of positive and negative samples. We address the problem of inseparability of distance distributions in existing unsupervised ReID models by introducing a global distance-distributions separation (GDS) constraint.

\item We maintain and update the distribution variables through a momentum update mechanism, enabling the timely update of the distribution variables and accurate estimation of the loss for each batch.



\item To further promote the separation of the Pos-distr and Neg-distr, we introduce a distribution-based hard mining mechanism in GDS.

\end{itemize}

GDS is simple yet effective and can be potentially used as a plug-and-play tool in many unsupervised ReID frameworks for performance enhancement. We validate the effectiveness of our GDS constraint in two representative UDA-ReID approaches, the clustering-based approach with pseudo labelling, and a style transfer-based approach. Extensive experiments are conducted on multiple ReID benchmark datasets and ours achieves the best performance.

\section{Related Work}

\subsection{Unsupervised Person Re-identification}

Unsupervised domain adaptive person ReID aims to learn a ReID model from a labeled source domain and an unlabeled target domain. It is attractive for real-word deployments as it does not require the expensive annotation efforts while exploiting the source domain knowledge. The domain gap between datasets results in poor generalization of a source domain trained model to another domain. Many domain adaptation techniques have been designed for person ReID. They can mainly be grouped into three categories: \emph{style transfer}, \emph{attribute recognition}, and \emph{clustering-based pseudo label estimation}. Style transfer based approaches translate the source labeled images to ones with the style of the target domain by using image style translation techniques (\egno, Cycle-GAN \cite{zhu2017unpaired}) for adaptation \cite{wei2018person,deng2018image,liu2019adaptive}. Their performance is much inferior to recent clustering-based approaches, since there is still a gap between the generated images and the target domain images \cite{yang2019asymmetric}. Attribute-based approaches \cite{wang2018transferable,lin2018multi} aim to share the source domain knowledge with the target domain through learning of some cues, such as person body attributes, to regularize the feature learning. Such external cues rely on manual annotation and thus limit their applicability. Clustering-based approaches are popular with superior performance \cite{song2020unsupervised,fan2018unsupervised,yang2019patch,zhang2019self,Fu2019SelfsimilarityGA,yang2019asymmetric}. The basic idea is to exploit the similarity of unlabeled samples by feature clustering for predicting pseudo labels and then use them for fine-tuning. Such clustering and training process are usually alternatively performed until the model is stable.

These unsupervised ReID models usually suffer from the interference from noisy pseudo labels or noisy generated images. The inseparability of the distance distributions of positive pairs and negative pairs is serious, there exists undesired small inter-class distances and large intra-class distances (see Fig. \ref{fig:motivation}(a)). 





\subsection{Metric Learning for Person Re-identification}

In person ReID tasks, metric learning aims to make the features of the same identity closer while pushing the features of different identities further apart \cite{hirzer2012relaxed,zhou2016deep,zheng2016person,wojke2018deep,ye2020deep}. Loss function designs play the role of metric learning to guide the feature representation learning. For person ReID, there are several widely studied loss functions with their variants. \emph{Identity loss} treats the training process of person ReID as an image classification problem in guiding the feature learning \cite{zheng2017person}. This enforces a distance/correlation constraint on samples with respect to the ``class center" but ignores the direct comparisons between samples. ReID inference is actually a comparison based on the feature similarity/distance between a query image and each image in the gallery set. \emph{Triplet-based loss} and its variants \cite{hermans2017defense} consider the relative ordering of the positive pair distance and negative pair distance by constructing a triplet (an anchor point $a$, a positive point $p$, a negative point $n$). They optimize the features to encourage the distance between the negative pair to be larger than the distance between the positive pair by a hard or soft margin. The relative ordering of positive pairs and negative pairs matches the retrieval purpose better. In recent years, triplet-based loss is popular and has become one of the default losses for most of the ReID networks \cite{zhang2019self,Fu2019SelfsimilarityGA,yang2019asymmetric,yang2019patch,ye2020deep}. However, it only jointly considers the separability among three samples and is far from satisfactory for the global comparison/ranking problem of ReID. Histogram loss \cite{ustinova2016learning} goes a step further and encourages the separability of the positive and negative distributions within a batch. As explained in Fig. \ref{fig:motivation2}, it is still a local constraint and results in poor separability. 

Besides capturing the local structure of data with triplet-based loss, Zhang \etal \cite{zhang2019self} use global information by appending a changeable classification layer to the model with the number of classes being the number of clusters. This encourages the separability of clustered centers but is less effective in encouraging the separation of distance distributions of positive pairs and negative pairs. Besides, the classification layer needs to be re-trained together with the change of clustering results which may result in a slow and unstable convergence.

To address the poor separability between the distance distributions of the positive pairs and negative pairs observed in existing unsupervised ReID approaches, we introduce a simple yet effective global distributions separation constraint with a momentum updating mechanism. The separation of global distributions has been investigated in learning local image descriptors, with applications in several problems involving the matching of local image patches, such as wide baseline stereo, structure from motion, image classification \cite{kumar2016learning}. Ours is superior to theirs. First, our momentum updating mechanism enables the timely and more accurate updating of losses and networks, leading to optimized performance. In contrast, theirs calculates the distribution statistics from all the samples to obtain the global loss before the next epoch (an epoch means that the entire dataset is passed). This loss is then used for all the mini-batches even though the actual statistics and losses vary with the optimization of each batch. Second, we propose a distribution-based hard mining for effectively promoting the separation. More importantly, we identify one key problem in unsupervised person ReID and address the challenge through a simple yet effective loss design and updating mechanism.

\begin{figure*}[t]
  \centerline{\includegraphics[width=0.95\linewidth]{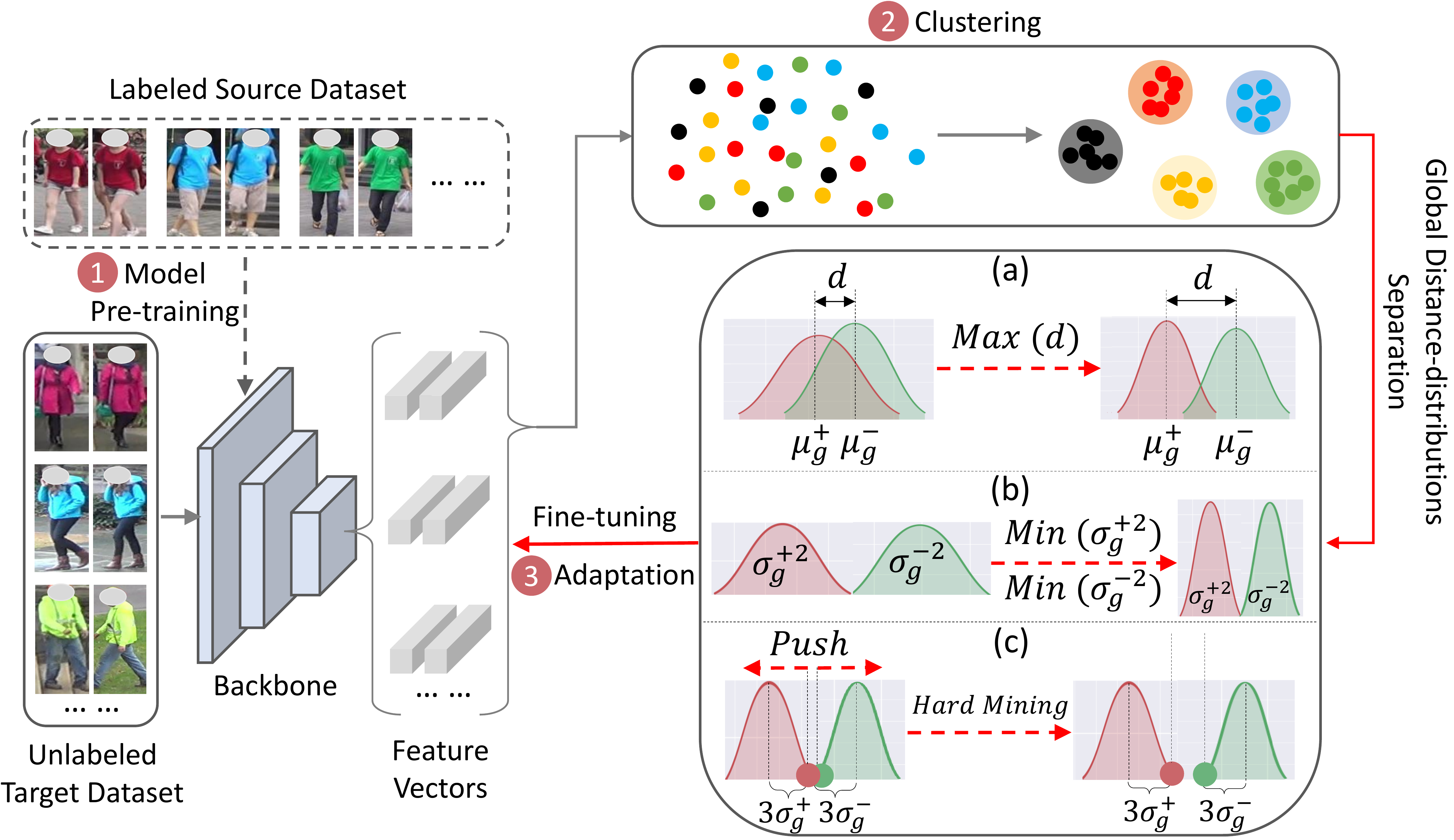}}
  \vspace{-1mm}
    \caption{Flowchart of the clustering-based UDA-ReID framework with our Global Distance-distributions Separation (GDS) constraint. In the third stage for adaptation, our GDS constraint with momentum updating optimizes the feature learning by encouraging the separation of the global Pos-distr (red curve) and Neg-distr (green curve). Particularly, we (a) enlarge the distance of the means of the two distributions; (b) encourage the sharpness of each distribution; and (c) introduce distribution-based hard mining to enforce a soft margin between the right-tail of the Pos-distr and the left-tail of the Neg-distr.}
\label{fig:pipeline}
\vspace{-6mm}
\end{figure*}

\section{Unsupervised ReID with GDS Constraint}

We aim at enhancing the performance of unsupervised person ReID, by addressing the inseparability of global distance-distributions of positive pairs and negative pairs. As illustrated in Fig. \ref{fig:motivation}, the inseparability could seriously degrade the ReID performance. We propose a global distance-distributions separation (GDS) constraint powered by a momentum update mechanism, which we also refer to as the GDS algorithm. Conceptually, the GDS algorithm is not limited to any specific unsupervised ReID network and it is general. To facilitate the better understanding of the GDS algorithm, we describe it under the representative and popular clustering-based framework for unsupervised person ReID.

Clustering-based approaches in general consist of three stages with the last two stages executed alternatively for many iterations \cite{song2020unsupervised,fan2018unsupervised,yang2019patch,zhang2019self,Fu2019SelfsimilarityGA,yang2019asymmetric}. As shown in Fig. \ref{fig:pipeline}, the first stage, \emph{model pre-training}, is to pretrain the ReID network with the source domain labeled data for feature learning. The second stage, \emph{clustering stage}, aims at assigning pseudo labels through the clustering results to the unlabeled target domain data. In the third stage of \emph{adaptation}, the pseudo labels are used for the fine-tuning of the network for domain adaptation, where triplet-based losses are often used for optimization. Triplet-based constraint is enforced on a triplet of instances and cannot assure the correct relative order of the distances for extensive pairs of samples (see Fig. \ref{fig:motivation}(a) for the illustration of the inseparability). We apply our GDS algorithm in the third stage for more effective feature learning. GDS encourages the separation of the global distance-distributions of positive sample pairs and negative sample pairs.

\subsection{Global Distance-distribution Modeling with Momentum Update}
\label{sec:GDM}

We have observed the dataset-wise distance distributions of positive sample pairs and negative sample pairs from ReID models (including unsupervised and supervised methods) and found they exhibit Gaussian-like distributions. Therefore, we model the distance distribution of sample pairs by Gaussian distribution with mean $\mu$ and variance $\sigma^2$. We denote the distance distribution of positive sample pairs as $\mathcal{N}(\mu^+, \sigma^{+2})$ and that of negative sample pairs as $\mathcal{N}(\mu^-, \sigma^{-2})$. We aim at designing optimization metrics and strategies to encourage the separation of the two dataset-wise (global) distributions in guiding the feature learning, which would benefit the distance-based ranking across all the images in the gallery set for ReID inference.

Instead of simultaneously exploring all samples, a convolutional neural network is optimized at the mini-batch level, where the loss calculated from the batch is back-propagated to optimize the network to facilitate timely update and avoid the requirement of large memory. Motivated by this, we maintain the dataset-wise distribution parameters and update them at every batch. This enables timely update of the estimated distribution variables and the corresponding GDS loss, where the distributions change with the updating of the feature extraction network. We formulate the update process here. Particularly, we maintain the global distance distribution of positive sample pairs with variables $\mu_{g}^+$ and $\sigma_{g}^{+2}$ and that of negative sample pairs with variables $\mu_{g}^-$ and $\sigma_{g}^{-2}$. We denote the local mean and variance corresponding to the distances of positive sample pairs within a local batch as $\mu_{l}^+$ and $\sigma_{l}^{+2}$ (and $\mu_{l}^-$ and $\sigma_{l}^{-2}$ for negative pairs). We formulate the momentum update of the global distance-distribution for the distance of positive pairs as
\begin{equation}
\left\{
\begin{array}{rl}
\mu_{g}^+& \leftarrow \beta \mu_{g}^+ + (1-\beta) \mu_{l}^+\\
\sigma_{g}^{+2}& \leftarrow \beta \sigma_{g}^{+2}+ (1-\beta) \sigma_{l}^{+2}
\end{array}
\begin{array}{rl}
{\rm, \hspace{0.2cm} with}
\end{array}
\begin{array}{rl}
\hspace{0.3cm}
\end{array}
\left\{
\begin{array}{rl}
\mu_{l}^+&=\frac{1}{N^+}\sum\nolimits_{i=1}^{N^+} d_i^+\\
\sigma_{l}^{+2}&=\frac{1}{N^+}\sum\nolimits_{i=1}^{N^+} (d_i^+ - \mu_g^+)^2
\end{array}
\right\}
\right\}
\label{update}
\end{equation}
where the hyper-parameter $\beta \in [0,1)$ is a momentum coefficient, $N^+$ denotes the number of sampled positive pairs. $d_i^+$ denotes the Euclidean distance of the $i^{th}$ positive pair, where $d(\mathbf{x}_{i_1},\mathbf{x}_{i_2}) = \frac{1}{2} ||\mathbf{x}_{i_1} - \mathbf{x}_{i_2}||\in [0,1]$ for the two normalized feature vectors $\mathbf{x}_{i_1}$ and $\mathbf{x}_{i_2}$ of two samples. The update for negative sample pairs is similar. Please see \textbf{Supplementary} for the mathematical analysis of the rationality of our momentum update design.

\subsection{Global Distance-distributions Separation (GDS) Constraint}

To encourage the separation of the global distance-distributions of positive sample pairs and negative sample pairs, we design a GDS loss with distribution-based hard mining, \ieno, GDS-H loss, which optimizes the network at each mini-batch. The GDS-H loss which consists of the basic GDS loss $\mathcal{L}_{GDS}$ and distribution-based hard mining loss $\mathcal{L}_{H}$ is defined as 

\begin{equation}
    \begin{aligned}
    \mathcal{L}_{GDS-H} & = \mathcal{L}_{GDS} + \lambda_h\mathcal{L}_{H},
    \end{aligned}
\label{eq:GDS-H}
\end{equation}
where $\lambda_h$ is a hyper-parameter that balances their importance. 

\noindent\textbf{GDS Loss.} $\mathcal{L}_{GDS}$ is defined as
\begin{equation}
    \begin{aligned}
    \mathcal{L}_{GDS} & = {\rm{Softplus}}(\mu_g^+ -\mu_g^-) + \lambda_{\sigma}(\sigma_{g}^{+2}+ \sigma_{g}^{-2}),
    \end{aligned}
\label{eq:GDS}
\end{equation}
where $\lambda_{\sigma}$ is another hyper-parameter that balances the importance of mean and variance (please see the study on $\lambda_h$ and $\lambda_{\sigma}$ in the experiment). $\rm{Softplus}(\cdot) = \ln{(1+\exp{(\cdot)})}$, as a soft-margin formulation, has similar behavior to the hinge function but it decays exponentially instead of having a hard cut-off \cite{hermans2017defense}. As illustrated in Fig. \ref{fig:pipeline}, the first item ${\rm{Softplus}}(\mu_g^+ -\mu_g^-)$ encourages the separation of the centers of the two distributions while the second item $(\sigma_{g}^{+2}+ \sigma_{g}^{-2})$ encourages the sharpness of the two distributions by minimizing their variances.


\noindent\textbf{Distribution-based Hard Mining.}
To better promote the separation of the two distributions, as illustrated in Fig. \ref{fig:pipeline}(c), we expect the hard samples lying in the potential overlap regions can also be separated. We achieve this by introducing a distribution-based hard mining loss $\mathcal{L}_{H}$ as
\begin{equation}
    \begin{aligned}
        \mathcal{L}_{H} & = {\rm{Softplus}}((\mu_g^+ + \kappa\sigma_{g}^+)-( \mu_g^- - \kappa\sigma_{g}^-)),
    \end{aligned}
    \label{eq:hardmining}
\end{equation}
where $\kappa$ is a hyper-parameter which controls the strength of hard mining. The larger of $\kappa$, the more areas of the distribution are covered. Motivated by the \emph{three-sigma rule of thumb} which expresses a conventional heuristic that nearly all values (99.7\%) are taken to lie within three times of standard deviations of the mean \cite{grafarend2012linear}, we set $\kappa$ to 3 and also experimentally found that this results in superior performance. As illustrated in Fig. \ref{fig:pipeline}(c), the physical meaning of Eq. (\ref{eq:hardmining}) is to encourage the right-tail of the Pos-distr to be smaller than the left-tail of the Neg-distr by a soft margin. This enables the relative ordering of the two \emph{distributions} (instead of only ordering of the centers of the distributions).

\noindent\textbf{Differentiability of GDS-H Loss for Optimization.}
Importantly, the GDS-H loss in Eq. (\ref{eq:GDS-H}) is differentiable w.r.t. the batch-level statistics $\mu_l$ and $\sigma_{l}^2$ and thus the batch samples. To facilitate the description, we denote the means of the maintained global distance-distribution of positive sample pairs as $\mu_g^+$ and $\mu_g^{+'}$ before and after the update using the current batch, respectively, where $\mu_g^{+'} = \beta \mu_{g}^+ + (1-\beta) \mu_{l}^+$. Similarly, we denote $\sigma_g^{+2'} = \beta \sigma_{g}^{+2}+ (1-\beta) \sigma_{l}^{+2}$. For ease of presentation, we rewrite the variables $\sigma_g^{+2'}$ and $\sigma_g^{+2}$ as $\rho_{g}^{+'}$ and $\rho_{g}^{+}$, respectively. Then $\rho_g^{+'} = \beta \rho_{g}^{+} + (1-\beta) \sigma_{l}^{+2}$. Similarly, we can define these for the global distance-distribution of negative sample pairs. The gradient of GDS-H loss ($\mathcal{L}_{GDS-H}=\mathcal{L}$) w.r.t. the sample feature $\mathbf{x}_{i_1}$ is as 
\begin{equation}
    \begin{aligned}
        \frac{\partial \mathcal{L}_{GDS-H}}{\partial\mathbf{x}_{i_1}} &= \frac{\partial \mathcal{L}}{\partial \mu_g^{+'}}\frac{\partial \mu_g^{+'}}{\partial \mu_l^{+}}\frac{\partial \mu_l^{+}}{\partial {d_i^+}}\frac{\partial d_i^+}{\partial{\mathbf{x}_{i_1}}} + \frac{\partial \mathcal{L}}{\partial \rho_g^{+'}}\frac{\partial \rho_g^{+'}}{\partial \sigma_l^+}\frac{\partial \sigma_l^+}{\partial{d_i^+}}\frac{\partial{d_i^+}}{\partial{\mathbf{x}_{i_1}}} \\
        &+ \frac{\partial \mathcal{L}}{\partial \mu_g^{-'}}\frac{\partial \mu_g^{-'}}{\partial \mu_l^{-}}\frac{\partial \mu_l^{-}}{\partial d_i^-}\frac{\partial{d_i^-}}{\partial{\mathbf{x}_{i_1}}} + \frac{\partial \mathcal{L}}{\partial \rho_g^{-'}}\frac{\partial \rho_g^{-'}}{\partial \sigma_l^-}\frac{\partial \sigma_l^-}{\partial{d_i^-}}\frac{\partial{d_i^-}}{\partial{\mathbf{x}_{i_1}}},
    \end{aligned}
\label{loss1_grad}
\end{equation}
this reveals that our method could enable the batch-level gradient back-propagation for timely update of the network for feature learning.

\section{Experiments}

For unsupervised person ReID, we describe the datasets and evaluation metrics in Subsection \ref{subsec:dataset}. Implementation details are introduced in Subsection \ref{subsec:implementation}. We perform ablation study and analysis in Subsection \ref{subsec:ablation}, which demonstrates the effectiveness of our GDS algorithm. The design choices for GDS are studied in Subsection \ref{subsec:design}. 
Subsection \ref{subsec:visualization} visualizes the dataset-wise (global) distance distributions. Subsection \ref{subsec:SOTA} shows the comparisons of our schemes with the state-of-the-art approaches. We apply our GDS constraint to two baseline networks and show significant gains in both cases.



\subsection{Datasets and Evaluation Metrics}
\label{subsec:dataset}

To evaluate the effectiveness of our GDS constraint for unsupervised person ReID and to be consistent with what were done in prior works
for performance comparisons, we conduct extensive experiments on the public ReID datasets, including 
Market1501 \cite{zheng2015scalable}, DukeMTMC-reID \cite{zheng2017unlabeled}, CUHK03 \cite{li2014deepreid}, and the large-scale MSMT17 \cite{wei2018person}. We denote Market1501 by M, DukeMTMC-reID by Duke or D, and CUHK03 by C for short. Given a labeled source dataset A and an unlabeled target dataset B, we denote this unsupervised ReID setting as A$\rightarrow $B.


We follow common practices and use the cumulative matching characteristics (CMC) at Rank-1, and mean average precision (mAP) for evaluation.

\subsection{Implementation Details}
\label{subsec:implementation}

We build our \emph{Baseline} following the clustering-based pseudo label approach \cite{song2020unsupervised}, which uses ResNet-50 \cite{almazan2018re,zhang2019DSA,luo2019bag,yang2019asymmetric,song2020unsupervised} as the backbone network for feature extraction. In the first stage, we pretrain the network using the source dataset, supervised by classification loss \cite{sun2018beyond,fu2019horizontal} and triplet loss with batch hard mining \cite{hermans2017defense}. In the clustering stage, we perform clustering using DBSCAN \cite{ester1996density} based on the extracted features of the target dataset, for the purpose of pseudo label assignment. In the adaptation stage, with the pseudo labels, triplet loss with batch hard mining \cite{hermans2017defense} is used to finetune the network. The second stage and the third stage are executed alternatively for 30 iterations. In our schemes, on top of \emph{Baseline}, we add the proposed GDS losses in the adaptation stage. The input image resolution is 256$\times$128. See \textbf{Supplementary} for more training details.  






\subsection{Ablation Study}
\label{subsec:ablation}

We perform most of the ablation studies based on the powerful clustering-based method, \ieno, \emph{Baseline}. We follow the popular settings \cite{Fu2019SelfsimilarityGA,yang2019asymmetric,zhong2019invariance} for ablation study, \ieno, using Market1501$\rightarrow$Duke (M$\rightarrow$D), and Duke$\rightarrow$Market1501 (D$\rightarrow$M).

\noindent\textbf{Effectiveness of Our GDS Constraint.} Table \ref{tab:ablations1}(a) shows the comparisons. \textbf{\emph{Direct transfer}} denotes the case that the network is only trained on source dataset and is directly tested on the target dataset. \textbf{\emph{Baseline+GDS}} denotes our basic scheme where our proposed GDS loss is added on top of \emph{Baseline}. \textbf{\emph{Baseline+GDS-H}} denotes our final scheme where our GDS-H loss (GDS loss + distribution-based hard-mining loss, with momentum update of distributions) is added on top of \emph{Baseline}. We have the following observations.

\begin{table*}[t]\centering
    \caption{Effectiveness of the proposed GDS loss and the distribution-based hard mining loss (H).}
    \vspace{1mm}
	\captionsetup[subfloat]{captionskip=0.5pt}
	\captionsetup[subffloat]{justification=centering}
	\subfloat[Study on the clustering-based method.\label{tab:Loss}]{
		\tablestyle{1.1pt}{1.2}
            \begin{tabular}{lcccc}
            \toprule
            \multicolumn{1}{c}{\multirow{2}[2]{*}{Clustering-based}} & \multicolumn{2}{c}{M$\rightarrow$D} & \multicolumn{2}{c}{D$\rightarrow$M} \\
                  & mAP   & Rank-1 & mAP   & Rank-1 \\
            \midrule
            Direct transfer & 19.8  & 35.3  & 21.8  & 48.3 \\
            Baseline  & 48.4  & 67.1  & 52.1  & 74.3 \\
            Baseline+\textbf{GDS} & {52.9} & {71.4} & {57.1} & {78.5} \\
            Baseline+\textbf{GDS-H} & {\textbf{55.1}} & {\textbf{73.1}} & {\textbf{61.2}} & {\textbf{81.1}} \\
            \bottomrule       
            \end{tabular}}
    \vspace{3mm}
	\subfloat[Study on style transfer based method. \label{tab:stage}]{
		\tablestyle{1.pt}{1.435}
        \begin{tabular}{lcccc}
        \toprule
        \multicolumn{1}{c}{\multirow{2}[2]{*}{Style Transfer based}} & \multicolumn{2}{c}{M$\rightarrow$D} & \multicolumn{2}{c}{D$\rightarrow$M} \\
              & mAP   & Rank-1 & mAP   & Rank-1 \\
        \midrule
 
        SPGAN \cite{deng2018image}  & 22.3  & 41.1  & 22.8  & 51.5 \\
        SPGAN+\textbf{GDS} & {25.3} & {45.9} & {25.9} & {56.4} \\
        SPGAN+\textbf{GDS-H} & {\textbf{27.5}} & {\textbf{47.7}} & {\textbf{29.4}} & {\textbf{60.8}} \\
        \bottomrule
        \end{tabular}%
		}
	\vspace{-11mm}
	\label{tab:ablations1}
\end{table*}

\begin{enumerate}[label=\arabic*),leftmargin=*,noitemsep,nolistsep]
\item Thanks to the encouragement of the separation of the global distance distributions, our final scheme \emph{Baseline+GDS-H} significantly outperforms \emph{Baseline} by \textbf{6.7\%} and \textbf{9.1\%} in mAP for M$\rightarrow$D and D$\rightarrow$M, respectively.  

\item Our distribution-based hard mining promotes the separation of the two distributions by enabling the relative ordering of the two distributions (instead of only their centers). Such hard mining further brings \textbf{2.2\%} and \textbf{4.1\%} improvements in mAP for M$\rightarrow$D and D$\rightarrow$M (\emph{Baseline+GDS-H} vs. \emph{Baseline+GDS}).

\item \emph{Baseline} which uses a clustering-based approach for adaptation outperforms \emph{Direct transfer} by a large margin, indicating the necessity of the adaptation with target domain data. 
\end{enumerate}

\begin{figure*}[t]
  \centerline{\includegraphics[width=0.9\linewidth]{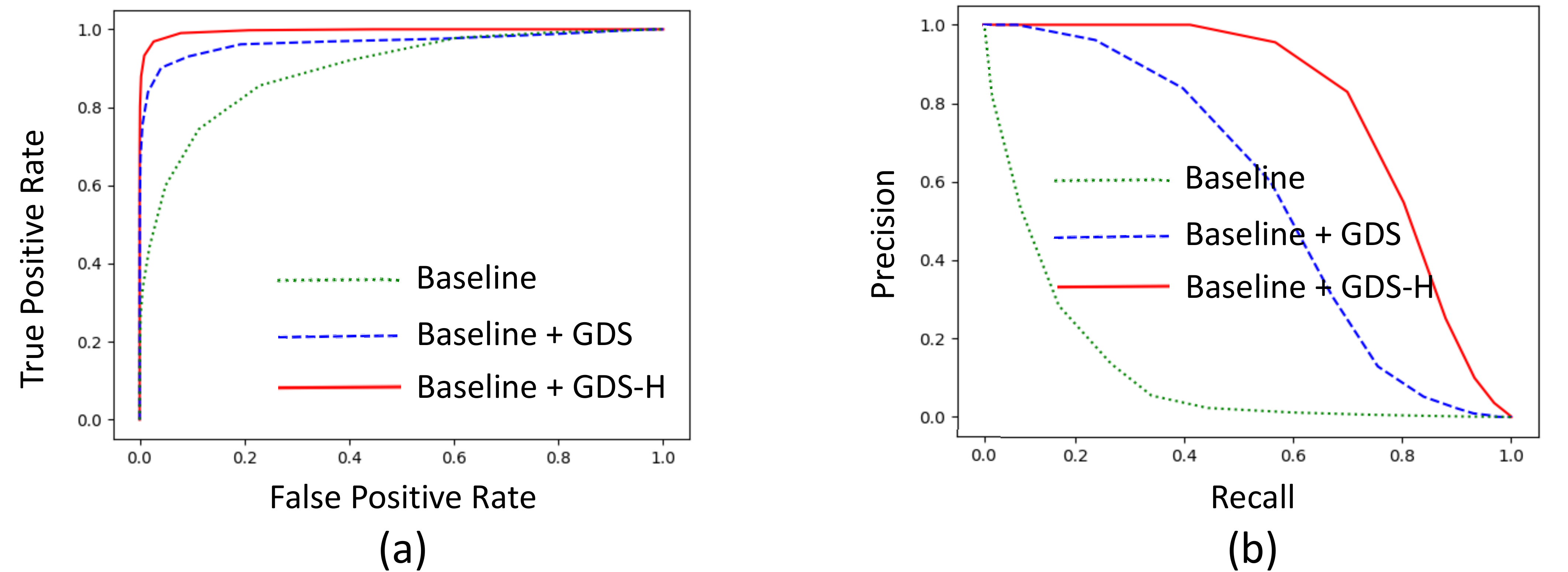}}
  \vspace{-2mm}
  \caption{(a) ROC curve and (b) PR curve on the test set of the target dataset Duke (Market1501$\rightarrow$Duke) for \emph{Baseline}, \emph{Baseline+GDS} and \emph{Baseline+GDS-H}.}
\label{fig:roc}
\vspace{-6mm}
\end{figure*}

\noindent\textbf{Effectiveness Evaluation in terms of ROC and PR Curves.} For some applications of ReID such as people tracking, the determination of whether two person images match or not degrades to a comparison of their feature distance with a pre-defined threshold $\theta$ \cite{ristani2018features,chen2018real}. If the distance is smaller than $\theta$, they are judged as the same person. By varying the threshould, we obtain the Receiver Operator Characteristic (ROC) curve \cite{fawcett2006introduction} and Precision-Recall (PR) curve \cite{davis2006relationship} as shown in Fig. \ref{fig:roc}. Obviously, our final scheme \emph{Baseline+GDS-H} also significantly outperforms \emph{Baseline} under the threshould-based evaluation metrics.

\noindent\textbf{Generality of our GDS Constraint.} The GDS constraint is not limited to clustering-based approaches and we also validate its effectiveness under a representative style transfer based ReID approach SPGAN~\cite{deng2018image}. Table \ref{tab:ablations1}(b) shows the results. We observe that \textbf{1)} our final scheme \emph{SPGAN+GDS-H} significantly outperforms \emph{SPGAN} by \textbf{5.2\%} and \textbf{6.6\%} in mAP for M$\rightarrow$D and D$\rightarrow$M, respectively; \textbf{2)} our distribution-based hard mining brings \textbf{2.2\%} and \textbf{3.5\%} improvements in mAP for M$\rightarrow$D and D$\rightarrow$M, respectively (\emph{SPGAN+GDS-H} vs. \emph{SPGAN+GDS}); \textbf{3)} this style transfer based approach \emph{SPGAN} is less effective than the clustering based approach \emph{Baseline} due to the unsatisfactory quality of the transferred images.


\begin{table}[t]
  \centering
  \scriptsize
  \caption{Performance comparison of different loss designs.}
    \renewcommand\arraystretch{1.2}
    \setlength{\tabcolsep}{3mm}{
    \begin{tabular}{lcccc}
    \toprule
    \multicolumn{1}{c}{\multirow{2}[2]{*}{Different losses}} & \multicolumn{2}{c}{M$\rightarrow$D} & \multicolumn{2}{c}{D$\rightarrow$M} \\
          & mAP   & Rank-1 & mAP   & Rank-1 \\
    \midrule
    Baseline & 48.4  & 67.1  & 52.1  & 74.3 \\
    Baseline + Histogram loss \cite{ustinova2016learning} & 50.2  & 69.4  & 53.7  & 76.3 \\
    Baseline + F-Statistic loss \cite{ridgeway2018learning} & 51.1  & 70.3  & 55.5  & 76.9 \\
    Baseline + Classification loss \cite{zhang2019self} & 49.7  & 68.9  & 53.8  & 75.9 \\
    Baseline + Global loss~\cite{kumar2016learning} & {49.6} & {68.8} & {54.3} & {76.5} \\
    \midrule

    

    Baseline + \textbf{GDS-H (Ours)} & \textbf{55.1} & \textbf{73.1} & \textbf{61.2} & \textbf{81.1} \\
    \bottomrule
    \end{tabular}}%
  \label{tab:losses}%
  \vspace{-1mm}
\end{table}%

\noindent\textbf{Comparison with Other Losses.}
We compare our GDS constraint with several other losses by implementing them on the same network \emph{Baseline} at its third stage (adaptation stage). 
Table \ref{tab:losses} shows the results. Both Histogram loss \cite{ustinova2016learning} and F-Statistic loss \cite{ridgeway2018learning} explore the local (batch-level) statistics for optimization. Histogram loss \cite{ustinova2016learning} estimates the similarity (or distance) distributions of positive sample pairs and negative sample pairs by accumulating the similarity (or distance) values to the bins of two histograms and minimizes their overlap. F-Statistic loss \cite{ridgeway2018learning} borrows a particular statistic from analysis of variance (ANOVA) hypothesis testing for equality of means. Adding a changeable fully connected layer followed by classification loss (with the number of classes equal to the number of clusters) \cite{zhang2019self} plays a role of  global constraint which encourages the separability of cluster centers. Global loss \cite{kumar2016learning} also encourages the global separation of distance distributions but there is a lack of timely update of distributions, resulting in inaccurate loss calculation and poor optimization.

We have the following observations: \textbf{1)} The performance of our proposed GDS-H loss significantly outperforms the other losses. \textbf{2)} With the local separation constraints, Histogram loss and F-Statistic loss both improve the performance over \emph{Baseline} but are inferior to our GDS-H loss. \textbf{3)} Adding a global classification loss brings about 1.3\%$\sim$1.7\% improvement in mAP over \emph{Baseline} but are not as effective as ours. \textbf{4)} Thanks to the effective momentum update mechanism and distribution-based hard mining, our GDS constraint significantly outperforms \cite{kumar2016learning}.

\subsection{Design Choices of GDS}
\label{subsec:design}

\begin{figure*}[t]
  \centerline{\includegraphics[width=1.0\linewidth]{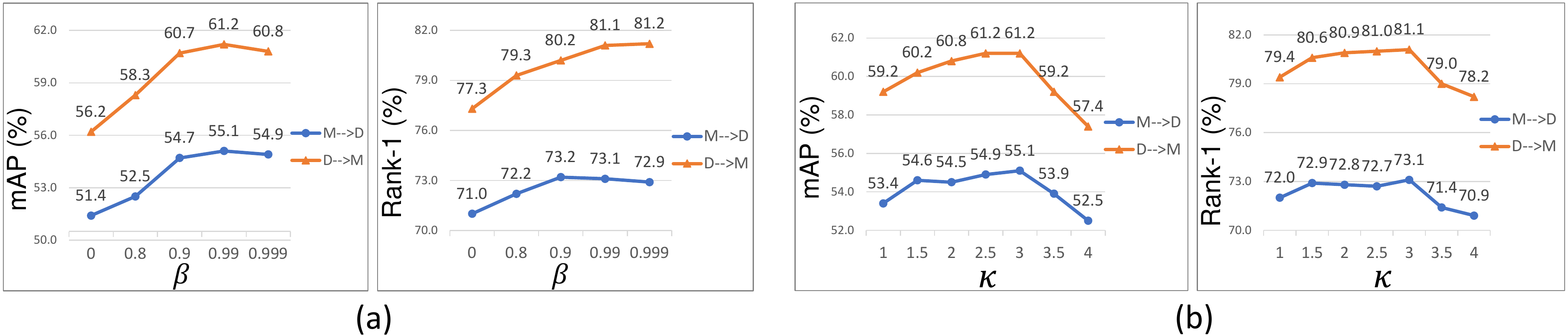}}
  \vspace{-2mm}
  \caption{Influence on ReID performance in terms of mAP and Rank-1 accuracy of (a) momentum coefficient $\beta$ (see Eq. (\ref{update}) in Subsection 3.1), and
  (b) strength $\kappa$ in the distribution-based hard mining for M$\rightarrow$D and D$\rightarrow$M settings.} 
   \label{fig:coef_and_str}
  \vspace{-4mm}
\end{figure*}

\noindent\textbf{Influence of the Momentum Coefficient $\beta$.} For the distribution update  in Subsection \ref{sec:GDM}), $\beta$ controls the contribution ratio of the batch-level statistics to the maintained distributions (see Eq. (\ref{update})). Fig. \ref{fig:coef_and_str}(a) shows its influence on ReID performance. We observe that a relatively larger value (\egno, $\beta = 0.99$) could help to maintain more stable global distributions and works much better than a smaller value. When $\beta$ is 0, our GDS loss degrades to a batch-wise (local) distance distribution separation constraint. Such degradation results in a large performance drop in comparison to our GDS ($\beta = 0.99$), from the best 55.1\%/61.2\% to 51.4\%/56.2\% in mAP for M$\rightarrow$D/D$\rightarrow$M, demonstrating the importance of modeling \emph{global} distance distributions.

\noindent\textbf{Influence of the strength $\kappa$ in the Distribution-based Hard Mining.} We compare the cases of adding different width offsets (from $\kappa=1$ to $4$) over the mean as the ``hard" region definition (see Eq. (\ref{eq:hardmining})) and show the performance in Fig. \ref{fig:coef_and_str}(b). We observe that $\kappa=3$ leads to the best performance. Based on \emph{three-sigma rule of thumb}, 99.7\% values lie within three times of standard deviations of the mean. This could well cover the entire distribution while excluding the side effects of some extreme outliers.




\noindent\textbf{Influence of the Hyper-parameters $\lambda_h$ and $\lambda_{\sigma}$.} We experimentally set $\lambda_{h}=0.5, \lambda_{\sigma}=1.0$ and please see \textbf{Supplementary} for more details.

\subsection{Visualization of Dataset-wise (Global) Distance Distributions}
\label{subsec:visualization}

\begin{figure*}[t]
  \centerline{\includegraphics[width=1.0\linewidth]{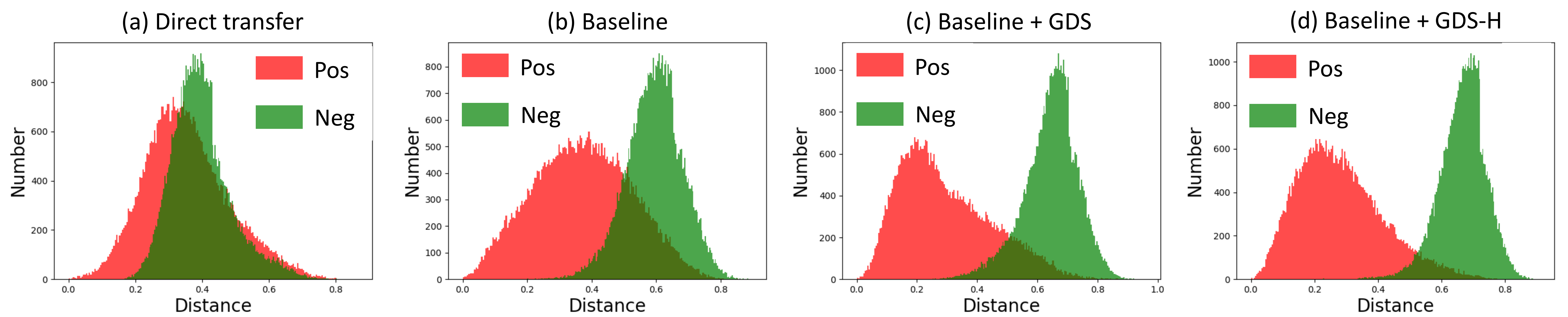}}
  \caption{Histograms of the distances of the positive sample pairs (red) and negative sample pairs (green) on the test set of the target dataset Duke (Market1501$\rightarrow$Duke) for schemes of (a) \emph{Direct transfer}, (b) \emph{Baseline}, (c) \emph{Baseline+GDS}, and (d) \emph{Baseline+GDS-H}. Here we use all the 48018 positive sample pairs and 48018 randomly sampled negative sample pairs for visualization.}
\label{fig:hist}
\vspace{-4mm}
\end{figure*}

To better understand how our GDS constraint works, we visualize the dataset-wise Pos-distr and Neg-distr in Fig. \ref{fig:hist} for four schemes.
Thanks to the adaptation on the unlabeled target dataset, the distance distributions of \emph{Baseline} present a much better separability than that of \emph{Direct transfer}. However, there is still a large overlap between the two distributions. By introducing our GDS constraint, our final scheme \emph{Baseline+GDS-H} greatly reduces the overlap of distributions. Besides, our distribution-based hard mining loss is very helpful in promoting the separation ((c) vs. (d)). 

\subsection{Comparison with State-of-the-Arts}
\label{subsec:SOTA}

Thanks to the capability of encouraging the separation of the two global distance distributions, our proposed GDS constraint effectively addresses the distance distributions inseparability problem observed in existing unsupervised ReID models. We evaluate the effectiveness of our GDS constraint by comparing with the state-of-the-art approaches on three datasets of Market-1501 (M), DukeMTMC-reID (D) and CUHK03 (C) with six settings in Table \ref{tab:STO}. More results about the largest dataset MSMT17 can be found in \textbf{Supplementary}.


With the effective loss design of GDS, our final scheme \emph{Baseline+GDS-H} achieves the best performance for five dataset settings without any increase in computation complexity in inference. Ours is inferior to \emph{PCB-R-PAST}* \cite{zhang2019self} only for the C$\rightarrow$D setting. But we should not look too much into this comparison as it is not a fair one. First, \emph{PCB-R-PAST}* applied re-ranking \cite{zhong2017re} as post-operation but we do not. Second, \emph{PCB-R-PAST}* is built on top of a more powerful ReID specific model structure of PCB \cite{sun2018beyond}. Ours uses ResNet-50. Third, the input resolution of \emph{PCB-R-PAST}* is 384$\times$128 while ours is 256$\times$128. Our \emph{Baseline+GDS-H} outperforms the second best approach \emph{ACT} \cite{yang2019asymmetric} on all the settings, achieving 4.6\% and 9.9\% gain in mAP for D$\rightarrow$C and C$\rightarrow$D, respectively. Besides, our loss design is simple in implementation and does not require the complicated co-training process used in \cite{yang2019asymmetric}. Conceptually, our GDS loss is complementary to many approaches like \emph{ACT}~\cite{yang2019asymmetric}, \emph{PCB-R-PAST}*~\cite{zhang2019self} and could be used to further improve their performance. 



To further demonstrate the effectiveness of our GDS constraint, we replace the ResNet-50 backbone network of \emph{Baseline} by a more powerful Style Normalization and Restitution (SNR) network \cite{jin2020snr} (which inserts four light-weight SNR modules to ResNet-50 to tackle the style variation problem for generalizable person ReID) and denote the new baseline scheme as \emph{B-SNR} for simplicity. In Table \ref{tab:STO}, the scheme \emph{B-SNR+GDS-H} which adopts our GDS constraint also significantly improves the performance of the strong baseline scheme \emph{B-SNR} and achieves the best performance on all settings.  

\begin{table}[t]
  \centering
  \tiny
  \caption{Performance (\%) comparisons with the state-of-the-art approaches for unsupervised person ReID. * means applying a re-ranking method of k-reciprocal encoding \cite{zhong2017re}. Note that \emph{Baseline} is built following \cite{song2020unsupervised} with ResNet-50 backbone and thus has nearly the same performance as \emph{Theory}~\cite{song2020unsupervised}. }
    \resizebox{1.0\textwidth}{!}{
    \begin{tabular}{lccccccccccccc}
    \toprule
    \multicolumn{1}{c}{\multirow{2}[2]{*}{Unsupervised ReID}} & \multirow{2}[2]{*}{Venue} & \multicolumn{2}{c}{M$\rightarrow$D} & \multicolumn{2}{c}{D$\rightarrow$M} & \multicolumn{2}{c}{M$\rightarrow$C} & \multicolumn{2}{c}{D$\rightarrow$C} & \multicolumn{2}{c}{C$\rightarrow$M} & \multicolumn{2}{c}{C$\rightarrow$D} \\
          &       & mAP   & Rank-1 & mAP   & Rank-1 & mAP   & Rank-1 & mAP   & Rank-1 & mAP   & Rank-1 & mAP   & Rank-1 \\
    \midrule
    PTGAN \cite{wei2018person} & CVPR'18 & --   & 27.4  & --   & 38.6  & --   & --   & --   & --   & --   & 31.5  & --   & 17.6 \\
    SPGAN \cite{deng2018image} & CVPR'18 & 22.3  & 41.1  & 22.8  & 51.5  & --   & --   & --   & --   & 19.0  & 42.8  & --   & -- \\
    TJ-AIDL \cite{wang2018transferable} & CVPR'18 & 23.0  & 44.3  & 26.5  & 58.2  & --   & --   & --   & --   & --   & --   & --   & -- \\
    HHL \cite{zhong2018generalizing} & ECCV'18 & 27.2  & 46.9  & 31.4  & 62.2  & --   & --   & --   & --   & 29.8  & 56.8  & 23.4  & 42.7 \\
    MAR \cite{yu2019unsupervised} & CVPR'19 & 48.0  & 67.1  & 40.0  & 67.7  & --   & --   & --   & --   & --   & --   & --   & -- \\
    ECN \cite{zhong2019invariance} & CVPR'19 & 40.4  & 63.3  & 43.0  & 75.1  & --   & --   & --   & --   & --   & --   & --   & -- \\
    PAUL \cite{yang2019patch} & CVPR'19 & 53.2  & 72.0  & 40.1  & 68.5  & --   & --   & --   & --   & --   & --   & --   & -- \\
    SSG \cite{Fu2019SelfsimilarityGA} & ICCV'19 & 53.4  & \textcolor[rgb]{ .267,  .447,  .769}{73.0} & 58.3  & 80.0  & --   & --   & --   & --   & --   & --   & --   & -- \\
    PCB-R-PAST$^*$ \cite{zhang2019self} & ICCV'19 & 54.3  & 72.4  & 54.6  & 78.4  & --   & --   & --   & --   & 57.3  & 79.5  & \textcolor[rgb]{ 1,  0,  0}{\textbf{51.8}} & \textcolor[rgb]{ 1,  0,  0}{\textbf{69.9}} \\
    Theory \cite{song2020unsupervised} & PR'2020 & 48.4  & 67.0  & 52.0  & 74.1  & 46.4  & 47.0  & 28.8  & 28.5  & 51.2  & 71.4  & 32.2  & 49.4 \\
    ACT \cite{yang2019asymmetric} & AAAI'20 & \textcolor[rgb]{ .267,  .447,  .769}{54.5} & 72.4  & \textcolor[rgb]{ .267,  .447,  .769}{60.6} & \textcolor[rgb]{ .267,  .447,  .769}{80.5} & \textcolor[rgb]{ .267,  .447,  .769}{48.9} & \textcolor[rgb]{ .267,  .447,  .769}{49.5} & \textcolor[rgb]{ .267,  .447,  .769}{30.0}  & \textcolor[rgb]{ .267,  .447,  .769}{30.6}  & \textcolor[rgb]{ .267,  .447,  .769}{64.1} & \textcolor[rgb]{ .267,  .447,  .769}{81.2}  & 35.4  & 52.8 \\
    \midrule
    Baseline & This work & 48.4  & 67.1  & 52.1  & 74.3  & 46.2  & 47.0  & 28.8  & 28.4  & 51.2  & 71.4  & 32.0  & 49.4 \\
    Baseline+\textbf{GDS-H}  & This work & \textcolor[rgb]{ 1,  0,  0}{\textbf{55.1}} & \textcolor[rgb]{ 1,  0,  0}{\textbf{73.1}} & \textcolor[rgb]{ 1,  0,  0}{\textbf{61.2}} & \textcolor[rgb]{ 1,  0,  0}{\textbf{81.1}} & \textcolor[rgb]{ 1,  0,  0}{\textbf{49.7}} & \textcolor[rgb]{ 1,  0,  0}{\textbf{50.2}} & \textcolor[rgb]{ 1,  0,  0}{\textbf{34.6}} & \textcolor[rgb]{ 1,  0,  0}{\textbf{36.0}} & \textcolor[rgb]{ 1,  0,  0}{\textbf{66.1}} & \textcolor[rgb]{ 1,  0,  0}{\textbf{84.2}} & \textcolor[rgb]{ .267,  .447,  .769}{45.3} & \textcolor[rgb]{ .267,  .447,  .769}{64.9} \\
    \midrule
    \midrule
    B-SNR\cite{jin2020snr} & CVPR'20 & 54.3	& 72.4	& 66.1	& 82.2	& 47.6	& 47.5	& 31.5	& 33.5	& 62.4	& 80.6	& 45.7	& 66.7 \\
    B-SNR\cite{jin2020snr}+\textbf{GDS-H}  & This work & {\textbf{59.7}}  & {\textbf{76.7}}  & {\textbf{72.5}}  & {\textbf{89.3}}  & {\textbf{50.7}}  & {\textbf{51.4}}  & {\textbf{38.9}}  & {\textbf{41.0}}  & {\textbf{68.3}}  & {\textbf{86.7}}  & {\textbf{51.0}}  & {\textbf{71.5}} \\
    \bottomrule
    \end{tabular}}
  \label{tab:STO}
  \vspace{-4mm}
\end{table}

\section{Conclusions}

For unsupervised person ReID, the serious inseparability of the distance distributions of the positive sample pairs and negative sample pairs significantly impacts the performance but is overlooked by the ReID research community. We propose the use of a Global Distance-distributions Separation (GDS) constraint to enhance the unsupervised person ReID performance. We model the global distance-distributions by Gaussian distributions and encourage their separation. Particularly, we exploit a momentum update mechanism to maintain the variables of the global distributions and enable the timely update of the distributions and the GDS-related loss, facilitating the optimization of the network for each batch. Moreover, we propose distribution-based hard mining to better promote the separation of the distributions. Extensive ablation studies demonstrate the effectiveness of our GDS constraint. We achieve the state-of-the-art performance on the bechmark datasets. The GDS design is simple yet effective. It is conceptually complementary to many of the available approaches and can be taken as a plug-and-play tool for ReID performance enhancement. 

\bibliographystyle{splncs04}
\bibliography{egbib}

\clearpage

\appendix
  \renewcommand\thesection{\arabic{section}}

\noindent{\LARGE \textbf{Supplementary}}
\vspace{5mm}

\section{Mathematical Analysis of the Rationality of Momentum Update Mechanism}

In order to mathematically prove the rationality of our momentum update design as described in Section 3.1 of the main manuscript, we design a toy game to present the momentum update process of the distance-distributions for different sample pair sets (\ieno, mean $\mu$, variance $\sigma^2$) with analysis/derivation.


Assume two random sets $\mathcal{A},\mathcal{B}$ with $N$ and $M$ sample pairs, respectively, and both sets exhibit Gaussian distribution. The mean and variance of set $\mathcal{A} = \{d_i|i=1,\cdots,N\}$ with $N$ sample pairs are represented as $\mu_A = \frac{1}{N} \sum_{i=1}^N d_i$, $\sigma^{2}_A = \frac{1}{N} \sum_{i=1}^N(d_i -\mu_A)^2$ while those of set $\mathcal{B} = \{d_j^{'}|j=1,\cdots,M\}$ with $M$ sample pairs are estimated by $\mu_B = \frac{1}{M} \sum_{j=1}^M {d_{j}^{'}}$, $\sigma^{2}_B = \frac{1}{M} \sum_{j=1}^M{(d_j^{'} - \mu_B)}^2$. We represent the set $\mathcal{C}$ as the combination of set $\mathcal{A}$ and set $\mathcal{B}$, the mean of the combined set $\mathcal{C}$ can be formulated as:
\begin{equation}
    \begin{aligned}
        \mu_{C} & = \frac{\sum_{i=1}^N d_i + \sum_{j=1}^M d_j^{'}}{N+M} = \frac{N}{N+M} \mu_A + \frac{M}{N+M} \mu_B = \beta \mu_A + (1-\beta) \mu_B,
    \end{aligned}
\label{update1}
\end{equation}
where $\beta = \frac{N}{N+M}$. Similarly, the variance of the combined set $\mathcal{C}$ can be obtained:
\begin{equation}
    \begin{aligned}
        \sigma^{2}_{C} & = \frac{\sum_{i=1}^N(d_i -\mu_{C})^2 + \sum_{j=1}^M(d_j^{'} -\mu_{C})^2}{N+M}, \\
    \end{aligned}
\label{update2}
\end{equation}
when $N$ is much larger than $M$ (just like the situation in our training where the number of the previously ``seen" mini-batches/samples is much larger than the number of samples in the current mini-batch), we could use $\mu_A$ to approximate $\mu_C$, \ieno, $\mu_C \approx \mu_A$, thus we can have:
\begin{equation}
    \begin{aligned}
        \sigma^{2}_{C} & \approx \frac{\sum_{i=1}^N(d_i -\mu_{A})^2 + \sum_{j=1}^M(d_j^{'} -\mu_{A})^2}{N+M} \\
        & = \frac{N}{N+M} \sigma^2_A + \frac{M}{N+M} \frac{\sum_{j=1}^M(d_j^{'} -\mu_{A})^2}{M} \\
        & = \beta \sigma^2_A + (1-\beta) \frac{\sum_{j=1}^M(d_j^{'} -\mu_{A})^2}{M}.
    \end{aligned}
\label{update2}
\end{equation}
By taking the sample pairs within a min-batch as the sample pairs of set $\mathcal{B}$, we can see that our momentum update design in Eq. (1) of our main manuscript is consistent with the above analysis/derivation.




\section{Details of Datasets}
\begin{table}[t]
  \centering
  \scriptsize
  \caption{Details about the ReID datasets.}
      \renewcommand\arraystretch{1.2}
    \setlength{\tabcolsep}{2mm}{
    \begin{tabular}{c|c|c|c|c|c}
    \toprule
    Datasets & Abbreviation & Identities & Images & Cameras & Scene \\
    \midrule
    Market1501 \cite{zheng2015scalable} & M & 1501  & 32668 & 6     & outdoor \\
    DukeMTMC-reID \cite{zheng2017unlabeled} & D & 1404  & 32948 & 8     & outdoor \\
    CUHK03 \cite{li2014deepreid} & C & 1467  & 28192 & 2     & indoor \\
    MSMT17 \cite{wei2018person} & MSMT17 & 4101  & 126142 & 15    & outdoor, indoor \\
    \bottomrule
    \end{tabular}}%
  \label{tab:datasets}%
\end{table}%

In Table \ref{tab:datasets}, we present the detailed information about the related person ReID datasets. Market1501 \cite{zheng2015scalable}, DukeMTMC-reID \cite{zheng2017unlabeled}, CUHK03 \cite{li2014deepreid}, and large-scale MSMT17 \cite{wei2018person} are the most commonly used datasets for unsupervised domain adaptive person ReID \cite{yu2019unsupervised,zhang2019self,Fu2019SelfsimilarityGA} and fully supervised person ReID \cite{zhang2019DSA,zhou2019omni}.
Market1501, DukeMTMC-reID, CUHK03, and MSMT17 all have commonly used pre-established train and test splits, which we use for our training and cross dataset test (\egno, M$\rightarrow$D, D$\rightarrow$M).

\section{Implementation Details}


\noindent\textbf{Data Augmentation and Training.} In the first stage of \emph{model pre-training}, just as in \cite{song2020unsupervised}, we use the commonly used data augmentation strategies of random cropping \cite{wang2018resource,zhang2019DSA}, horizontal flipping, random erasing (REA) \cite{luo2019bag,zhou2019omni}, and the label smoothing regularization \cite{szegedy2016rethinking} to train the network for obtaining the capability of extracting discriminative features for person ReID on the labeled source dataset. The training is supervised by classification loss \cite{sun2018beyond,fu2019horizontal} and triplet loss with batch hard mining \cite{hermans2017defense}. In the second stage of \emph{clustering}, we discard all the previous data augmentation operations and just simply extract features for the images of the target datasets for clustering. For the third stage of \emph{adaptation}, consistent with the operations in the first stage, we leverage all these data augmentations to fine-tune the network. 




In the first and third stages, following \cite{hermans2017defense}, a batch is formed by first randomly sampling $P$ identities. For each identity, we sample $K$ images. Then the batch size is $B=P\times K$. We set $P=32$ and $K=4$ (\ieno, batch size $B=P\times K=128$). We use Adam optimizer \cite{kingma2014adam} for both stages. 

For the first stage of \emph{model pre-training}, we set the initial learning rate to 3$\times$10$^{-4}$ and regularize the network with a weight decay of 5$\times$10$^{-4}$. The learning rate is decayed by a factor of 0.1 for every 50 epochs. We train the model on the source dataset for a total of 150 epochs. For the third  stage of \emph{adaptation}, we set the  learning rate to 6$\times$10$^{-5}$ and keep it unchanged. The second stage and the third  stage are executed alternatively for 30 iterations. For each iteration, we train our model for 70 epochs (that means, traverse all the target training samples for 70 times). For our proposed schemes, on top of \emph{Baseline}, we add the proposed GDS constraint in the third stage.

All our models are implemented on PyTorch and trained on a single 16G NVIDIA-P100 GPU. We will release our code upon acceptance.

\begin{table}[t]
  \centering
  \caption{Performance (\%) comparisons with the state-of-the-art approaches for unsupervised person ReID on the target dataset MSMT17 \cite{wei2018person}.}
  \setlength{\tabcolsep}{3mm}{
    \begin{tabular}{lccccc}
    \toprule
    \multicolumn{1}{c}{\multirow{2}[2]{*}{Unsupervised ReID}} & \multirow{2}[2]{*}{Venue} & \multicolumn{2}{c}{M$\rightarrow$MSMT17} & \multicolumn{2}{c}{D$\rightarrow$MSMT17} \\
          &       & mAP   & Rank-1 & mAP   & Rank-1 \\
    \midrule
    PTGAN \cite{wei2018person} & CVPR'18 & 2.9   & 10.2  & 3.3   & 11.8 \\
    SSG \cite{Fu2019SelfsimilarityGA} & ICCV'19 & \textcolor[rgb]{ .267,  .447,  .769}{13.2} & \textcolor[rgb]{ .267,  .447,  .769}{31.6} & \textcolor[rgb]{ .267,  .447,  .769}{13.3} & \textcolor[rgb]{ .267,  .447,  .769}{32.2} \\
    \midrule
    Baseline & This work & 7.2   & 18.9  & 9.2   & 25.3 \\

    Baseline+\textbf{GDS-H} & This work & \textcolor[rgb]{ 1,  0,  0}{\textbf{14.9}} & \textcolor[rgb]{ 1,  0,  0}{\textbf{34.3}} & \textcolor[rgb]{ 1,  0,  0}{\textbf{14.2}} & \textcolor[rgb]{ 1,  0,  0}{\textbf{33.9}} \\
    \bottomrule
    \end{tabular}}%
  \label{tab:msmt17}%
\end{table}%


\section{Influence of the Hyper-parameters $\lambda_h$ and $\lambda_{\sigma}$} 
The hyper-parameter $\lambda_h$ is used to balance the importance between the basic GDS loss $\mathcal{L}_{GDS}$ and the distribution-based hard mining loss $\mathcal{L}_{H}$. $\lambda_{\sigma}$ aims to balance the mean and variance constraints within $\mathcal{L}_{GDS}$. For $\lambda_h$ and $\lambda_{\sigma}$, we initially set them to 1, and then coarsely determine each one based on the corresponding loss values and their gradients observed during the training. The decision principle is to set their values to make the loss values/gradients lie in a similar range. Grid search within a small range of the derived $\lambda_h$/$\lambda_{\sigma}$ is further employed to get better parameters. Actually, we observed the final performance is not very sensitive to the two hyper-parameters, we experimentally set $\lambda_{h}=0.5, \lambda_{\sigma}=1.0$ in the end.

\section{Comparison with State-of-the-Arts (Complete Version)}
\label{subsec:SOTA}

More comparison results with state-of-the-art methods on the target dataset MSMT17 can be found in Table \ref{tab:msmt17}. We observe that in comparison with \emph{Baseline}, our GDS constraint brings gains of \textbf{7.7\%/15.4\%} and \textbf{5.0\%/8.6\%} in mAP/Rank-1 for M$\rightarrow$MSMT17 and D$\rightarrow$MSMT17, respectively, which demonstrates the effectiveness of our proposed GDS constraint. SSG \cite{Fu2019SelfsimilarityGA} also belongs to clustering-based approach. It exploits the potential similarity from the global body to local parts to build multiple clusters at different granularities. As a comparison, our \emph{Baseline} and \emph{Baseline+GDS-H} only consider the similarity at global body. Being simple in design, our final scheme \emph{Baseline+GDS-H} outperforms the second best method SSG \cite{Fu2019SelfsimilarityGA} by \textbf{2.7\%} and \textbf{1.7\%} in Rank-1 accuracy for M$\rightarrow$MSMT17 and D$\rightarrow$MSMT17, respectively. 

In addition, to save space, we only present the latest approaches in the Section 4.6 ``Comparison with State-of-the-Arts'' in the main manuscripts and here we show comparisons with more approaches in Table \ref{tab:STO}.

\begin{table}[t]
  \centering
  \tiny
  \caption{Performance (\%) comparisons with the state-of-the-art approaches for unsupervised person ReID. * means applying a re-ranking method of k-reciprocal encoding \cite{zhong2017re}. Note that \emph{Baseline} is built following \cite{song2020unsupervised} with ResNet-50 backbone and thus has nearly the same performance as \emph{Theory}\cite{song2020unsupervised}. To save space, we only present the latest approaches in the main manuscripts and here we show comparisons with more approaches.}
    \resizebox{1.0\textwidth}{!}{
    \begin{tabular}{lccccccccccccc}
    \toprule
    \multicolumn{1}{c}{\multirow{2}[2]{*}{Unsupervised ReID}} & \multirow{2}[2]{*}{Venue} & \multicolumn{2}{c}{M$\rightarrow$D} & \multicolumn{2}{c}{D$\rightarrow$M} & \multicolumn{2}{c}{M$\rightarrow$C} & \multicolumn{2}{c}{D$\rightarrow$C} & \multicolumn{2}{c}{C$\rightarrow$M} & \multicolumn{2}{c}{C$\rightarrow$D} \\
          &       & mAP   & Rank-1 & mAP   & Rank-1 & mAP   & Rank-1 & mAP   & Rank-1 & mAP   & Rank-1 & mAP   & Rank-1 \\
    \midrule
    CAMEL \cite{yu2017cross} & ICCV'17 & --   & --   & 26.3  & 54.5  & --   & --   & --   & --   & --   & --   & --   & -- \\
    PUL \cite{fan2018unsupervised} & TOMM'18 & --   & --   & 20.5  & 45.5  & --   & --   & --   & --   & --   & --   & --   & -- \\
    PTGAN \cite{wei2018person} & CVPR'18 & --   & 27.4  & --   & 38.6  & --   & --   & --   & --   & --   & 31.5  & --   & 17.6 \\
    SPGAN \cite{deng2018image} & CVPR'18 & 22.3  & 41.1  & 22.8  & 51.5  & --   & --   & --   & --   & 19.0  & 42.8  & --   & -- \\
    TJ-AIDL \cite{wang2018transferable} & CVPR'18 & 23.0  & 44.3  & 26.5  & 58.2  & --   & --   & --   & --   & --   & --   & --   & -- \\
    ARN \cite{li2018adaptation} & CVPRW'18 & 33.4  & 60.2  & 39.4  & 70.3  & --   & --   & --   & --   & --   & --   & --   & -- \\
    MMFA \cite{lin2018multi} & BMVC'18 & 24.7  & 45.3  & 27.4  & 56.7  & --   & --   & --   & --   & --   & --   & --   & -- \\
    HHL \cite{zhong2018generalizing} & ECCV'18 & 27.2  & 46.9  & 31.4  & 62.2  & --   & --   & --   & --   & 29.8  & 56.8  & 23.4  & 42.7 \\
    CFSM \cite{chang2019disjoint} & AAAI'19 & 27.3  & 49.8  & 28.3  & 61.2  & --   & --   & --   & --   & --   & --   & --   & -- \\
    MAR \cite{yu2019unsupervised} & CVPR'19 & 48.0  & 67.1  & 40.0  & 67.7  & --   & --   & --   & --   & --   & --   & --   & -- \\
    ECN \cite{zhong2019invariance} & CVPR'19 & 40.4  & 63.3  & 43.0  & 75.1  & --   & --   & --   & --   & --   & --   & --   & -- \\
    PAUL \cite{yang2019patch} & CVPR'19 & 53.2  & 72.0  & 40.1  & 68.5  & --   & --   & --   & --   & --   & --   & --   & -- \\
    SSG \cite{Fu2019SelfsimilarityGA} & ICCV'19 & 53.4  & \textcolor[rgb]{ .267,  .447,  .769}{73.0} & 58.3  & 80.0  & --   & --   & --   & --   & --   & --   & --   & -- \\
    PCB-R-PAST$^*$ \cite{zhang2019self} & ICCV'19 & 54.3  & 72.4  & 54.6  & 78.4  & --   & --   & --   & --   & 57.3  & 79.5  & \textcolor[rgb]{ 1,  0,  0}{\textbf{51.8}} & \textcolor[rgb]{ 1,  0,  0}{\textbf{69.9}} \\
    Theory \cite{song2020unsupervised} & PR'2020 & 48.4  & 67.0  & 52.0  & 74.1  & 46.4  & 47.0  & 28.8  & 28.5  & 51.2  & 71.4  & 32.2  & 49.4 \\
    ACT \cite{yang2019asymmetric} & AAAI'20 & \textcolor[rgb]{ .267,  .447,  .769}{54.5} & 72.4  & \textcolor[rgb]{ .267,  .447,  .769}{60.6} & \textcolor[rgb]{ .267,  .447,  .769}{80.5} & \textcolor[rgb]{ .267,  .447,  .769}{48.9} & \textcolor[rgb]{ .267,  .447,  .769}{49.5} & \textcolor[rgb]{ .267,  .447,  .769}{30.0}  & \textcolor[rgb]{ .267,  .447,  .769}{30.6}  & \textcolor[rgb]{ .267,  .447,  .769}{64.1} & \textcolor[rgb]{ .267,  .447,  .769}{81.2}  & 35.4  & 52.8 \\
    \midrule
    Baseline & This work & 48.4  & 67.1  & 52.1  & 74.3  & 46.2  & 47.0  & 28.8  & 28.4  & 51.2  & 71.4  & 32.0  & 49.4 \\
    Baseline + GDS & This work & 52.9  & 71.4  & 57.1  & 78.5  & 48.0  & 48.9  & 30.7 & 32.5 & 63.6  & 81.6 & 44.1  & 64.0 \\
    Baseline + GDS-H  & This work & \textcolor[rgb]{ 1,  0,  0}{\textbf{55.1}} & \textcolor[rgb]{ 1,  0,  0}{\textbf{73.1}} & \textcolor[rgb]{ 1,  0,  0}{\textbf{61.2}} & \textcolor[rgb]{ 1,  0,  0}{\textbf{81.1}} & \textcolor[rgb]{ 1,  0,  0}{\textbf{49.7}} & \textcolor[rgb]{ 1,  0,  0}{\textbf{50.2}} & \textcolor[rgb]{ 1,  0,  0}{\textbf{34.6}} & \textcolor[rgb]{ 1,  0,  0}{\textbf{36.0}} & \textcolor[rgb]{ 1,  0,  0}{\textbf{66.1}} & \textcolor[rgb]{ 1,  0,  0}{\textbf{84.2}} & \textcolor[rgb]{ .267,  .447,  .769}{45.3} & \textcolor[rgb]{ .267,  .447,  .769}{64.9} \\
    \midrule
    \midrule
    B-SNR\cite{jin2020snr} & CVPR'20 & 54.3	& 72.4	& 66.1	& 82.2	& 47.6	& 47.5	& 31.5	& 33.5	& 62.4	& 80.6	& 45.7	& 66.7 \\
    B-SNR\cite{jin2020snr}+GDS & This work & 57.2  & 74.6  & 68.6  & 84.9  & 49.8  & 50.5  & 36.7  & 38.8  & 67.2  & 85.1  & 49.4  & 69.9 \\
    B-SNR\cite{jin2020snr}+GDS-H  & This work & {\textbf{59.7}}  & {\textbf{76.7}}  & {\textbf{72.5}}  & {\textbf{89.3}}  & {\textbf{50.7}}  & {\textbf{51.4}}  & {\textbf{38.9}}  & {\textbf{41.0}}  & {\textbf{68.3}}  & {\textbf{86.7}}  & {\textbf{51.0}}  & {\textbf{71.5}} \\
    \bottomrule
    \end{tabular}}%
  \label{tab:STO}%
  \vspace{-2mm}
\end{table}%

\begin{figure}[t]
  \centerline{\includegraphics[width=1.0\linewidth]{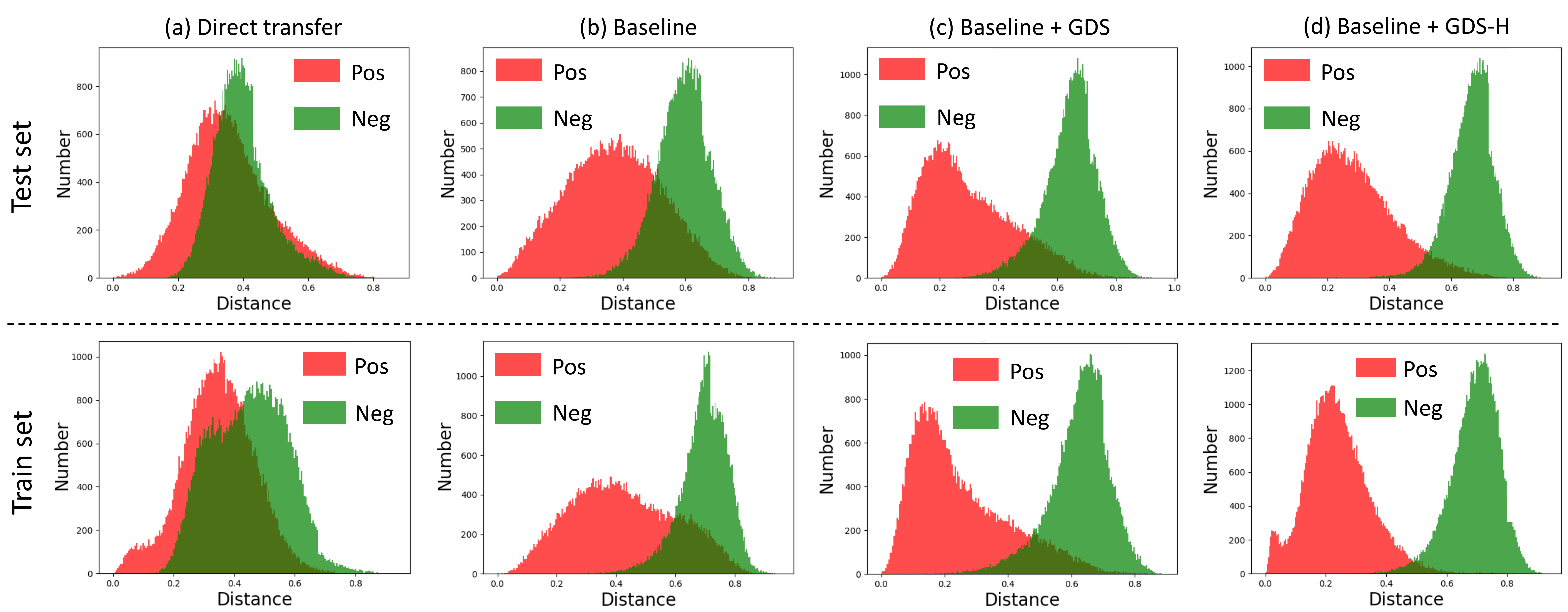}}
  \caption{Histograms of the distances of the positive sample pairs (red) and negative sample pairs (green) on the \textbf{test set (top)} and \textbf{train set (bottom)} of the target dataset Duke (Market1501$\rightarrow$Duke) for schemes of (a) \emph{Direct transfer}, (b) \emph{Baseline}, (c) \emph{Baseline+GDS}, and (d) \emph{Baseline+GDS-H}.}
\label{fig:hist2}
\end{figure}

\begin{figure}[t]
  \centerline{\includegraphics[width=1.0\linewidth]{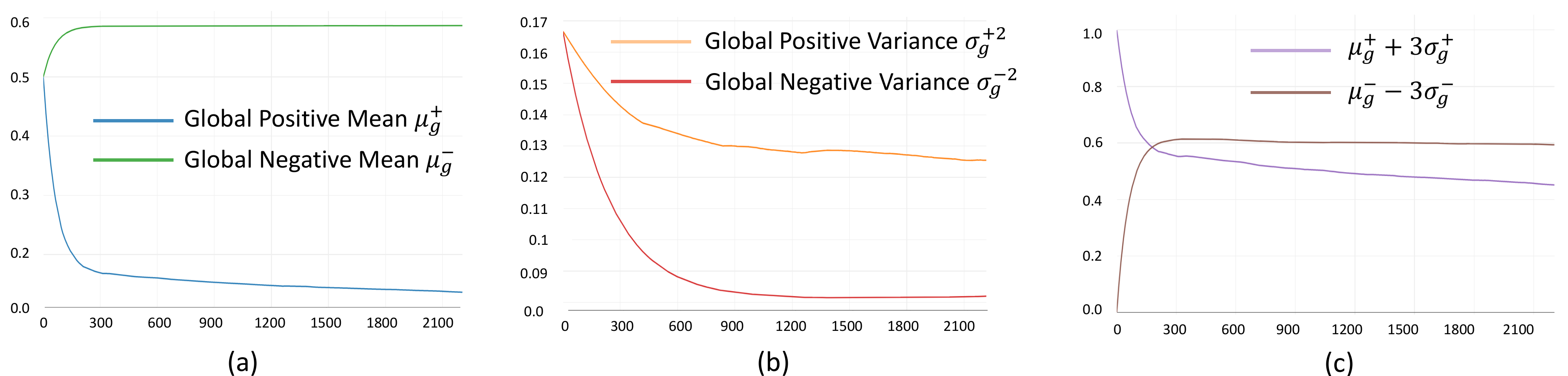}}
  \caption{Trend analysis of the learned dataset-wise (global) statistics in the training.}
\label{fig:trend}
\end{figure}

\section{More Visualization Results}

\noindent\textbf{Visualization of Dataset-wise (Global) Distance Distributions.} To better understand how well our GDS constraint works, in Fig. \ref{fig:hist2}, we not only visualize the dataset-wise Pos-distr and Neg-distr on the test set of target dataset (as shown in Fig.~6 in the main manuscripts), but also visualize the counterpart on the training set of target dataset. We have the following observations. 1) Thanks to the adaptation on the unlabeled target dataset and our GDS constraint, the distance distributions of our final scheme \emph{Baseline+GDS-H} present a much better separability than that of other schemes. This trend can be observed on both the training set and test set. 2) On the training set, each scheme presents better separability than that on the test set, especially for our final scheme\emph{Baseline+GDS-H}, which suggests that our GDS constraint is actually very helpful in promoting the separation after the optimization.

\noindent\textbf{Trend Analysis of the Learned Dataset-wise (Global) Statistics.} We observe the changing trend of the global statistics of distance distributions (including the mean $\mu_{g}^+$ of global Pos-distr, the mean $\mu_{g}^-$ of global Neg-distr, the variance $\sigma_{g}^{+2}$ of global Pos-distr, and the variance $\sigma_{g}^{-2}$ of global Neg-distr) in the training process and show the curves in Fig.~\ref{fig:trend}\footnote{We initialize the two distributions with mean of 0.5 and variance of 1/6 for the observation. Actually, we found the performance is not sensitive to the initialization values of the statistics.}. The horizontal axis denotes the identities of the epochs (30 iterations $\times$ 70 epochs = 2100 epochs). We observe that 1) as we expected, the centers/means of two distributions ($\mu_{g}^+, \mu_{g}^-$) and their hard tails ($\mu_{g}^{+}+3\sigma_{g}^{+}, \mu_{g}^{-}-3\sigma_{g}^{-}$) become further apart as the training goes; 2) the two distributions variance ($\sigma_{g}^{+2},\sigma_{g}^{-2}$) become sharper since the variances become smaller as the training progresses.    


\section{GDS Constraint Applied to Supervised Person ReID}

\begin{figure*}[t]
  \centerline{\includegraphics[width=0.45\linewidth]{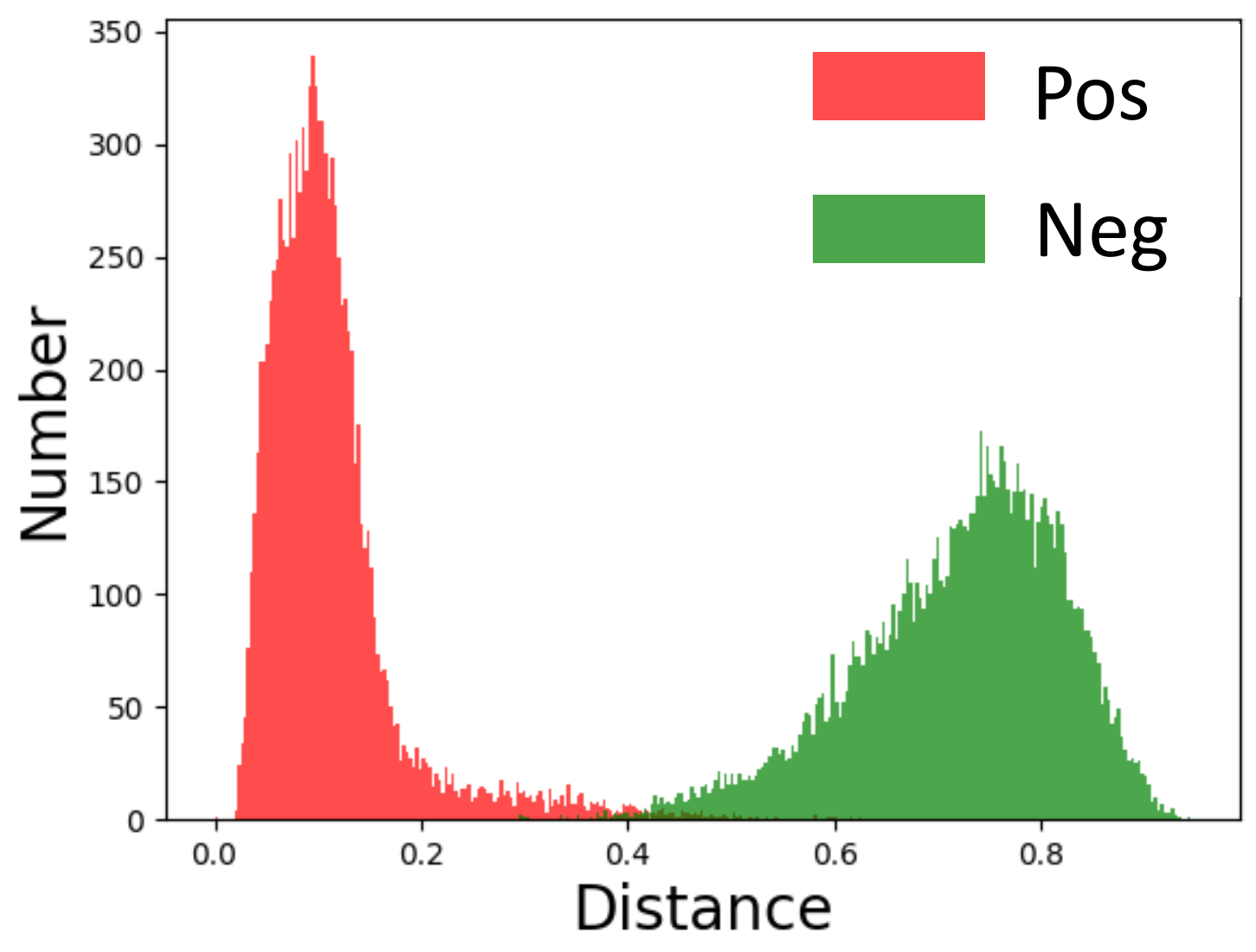}}
  \caption{Histograms of the distances of the positive sample pairs (red) and negative sample pairs (green) on the \emph{training} set of the labeled dataset CUHK03 for fully supervised person ReID.}
\label{fig:hist-supp}
\end{figure*}

We design the GDS constraint for addressing the inseparability of distance distributions in unsupervised person ReID, where there is no groudtruth labels for the target dataset. The use of either the pseudo labels or style transferred images results in noises and overlapping of the two distributions. For fully supervised person ReID, the proposed GDS is also expected to enhance the performance. However, on the benchmark datasets, due to the use of reliable labels and the over-fitting problem, we found the distance distributions on the training set are already well separated (see Fig. \ref{fig:hist-supp}) and thus there left small optimization space for us. Quantitatively, as shown in Table \ref{tab:full}, although our GDS brings some performance improvement (1.6\% and 1.9\% in mAP for CUHK03(L) and MSMT17, respectively), it is not significant in comparison with the unsupervised ReID setting.

\begin{table}[t]
  \centering
  \caption{Effectiveness of the proposed GDS loss and the distribution-based hardmining loss (H) for the fully supervised Person ReID.}
  \setlength{\tabcolsep}{3mm}{
    \begin{tabular}{lcccc}
    \toprule
    \multicolumn{1}{c}{\multirow{2}[2]{*}{Supervised ReID}} & \multicolumn{2}{c}{CUHK03 (L)} & \multicolumn{2}{c}{MSMT17} \\
          & mAP   & Rank-1 & mAP   & Rank-1 \\
    \midrule
    Baseline & 69.8  & 73.7  & 47.2  & 73.8 \\
    Baseline+\textbf{GDS} & 70.7  & 74.3  & 48.3  & 74.4 \\
    Baseline+\textbf{GDS-H} & \textbf{71.4}  & \textbf{75.5}  & \textbf{49.1}  & \textbf{74.9} \\
    \bottomrule
    \end{tabular}}%
  \vspace{-2mm}
  \label{tab:full}%
\end{table}%

\section{Training Complexity Analysis}

The increase of training time of our design in comparison with \emph{Baseline} \cite{song2020unsupervised} is negligible. We build \emph{Baseline} with the representative clustering-based method \cite{song2020unsupervised}, and add the proposed GDS constraint in the training. Both our loss calculation and momentum update have very low computation complexity in comparison with the convolutional operations of the network. Take the setting of using DukeMTMC-reID as source dataset and Market1501 as target dataset as an example, the training time of \emph{Baseline} \cite{song2020unsupervised} and our scheme \emph{Baseline+GDS-H} is 17.9 hours and 18.2 hours, respectively (\ieno, about 1.7\% increase). The training time is comparable to that of the existing STOA methods (PAUL \cite{yang2019patch} with 16.3 hours, SSG \cite{Fu2019SelfsimilarityGA} with 20.8 hours, MAR \cite{yu2019unsupervised} with 25.6 hours). Note that all these training courses are conducted on a single 16G NVIDIA-P100 GPU.

\begin{table}[t]
  \centering
  \caption{Performance w.r.t different clustering algorithms for the M$\rightarrow$D setting.}
  \setlength{\tabcolsep}{3mm}{
    \begin{tabular}{lcccccc}
    \toprule
    \multicolumn{1}{c}{\multirow{2}[2]{*}{M$\rightarrow$D}} & \multicolumn{2}{c}{K-means} &
    \multicolumn{2}{c}{DBSCAN} & \multicolumn{2}{c}{HDBSCAN} \\
          & mAP   & Rank-1 & mAP   & Rank-1 & mAP   & Rank-1\\
    \midrule
    Baseline & 40.2  & 57.9  & 48.4  & 67.1 & 49.6 & 67.9\\
    Baseline+\textbf{GDS-H} & 49.0  & 67.2  & 55.1  & 73.1  & 55.7 & 73.6 \\
    \bottomrule
    \end{tabular}}%
  \vspace{-2mm}
  \label{tab:full}%
\end{table}%

\section{Performance w.r.t Different Clustering Algorithms}

The performance of the \emph{Baseline} scheme with the cluttering approach of DBSCAN is similar to that with hierarchical DBSCAN (HDBSCAN), 48.4\% vs. 49.6\% in mAP for M$\rightarrow$D, and both outperforms the \emph{Baseline} scheme with K-means (40.2\%). Our GDS constraint consistently brings improvement of 8.8\%, 6.7\%, and 6.1\% for that with K-means, DBSCAN, and HBSCAN, respectively. For simplicity, we use DBSCAN by default in our experiments.

%
%
\bibliographystyle{splncs04}
\bibliography{egbib}

\begin{thebibliography}{10}
\providecommand{\url}[1]{\texttt{#1}}
\providecommand{\urlprefix}{URL }
\providecommand{\doi}[1]{https://doi.org/#1}

\bibitem{almazan2018re}
Almazan, J., Gajic, B., Murray, N., Larlus, D.: Re-id done right: towards good
  practices for person re-identification. arXiv preprint arXiv:1801.05339
  (2018)

\bibitem{chang2019disjoint}
Chang, X., Yang, Y., Xiang, T., Hospedales, T.M.: Disjoint label space transfer
  learning with common factorised space. In: AAAI. vol.~33, pp. 3288--3295
  (2019)

\bibitem{chen2018real}
Chen, L., Ai, H., Zhuang, Z., Shang, C.: Real-time multiple people tracking
  with deeply learned candidate selection and person re-identification. In:
  ICME. pp.~1--6 (2018)

\bibitem{davis2006relationship}
Davis, J., Goadrich, M.: The relationship between precision-recall and roc
  curves. In: ICML. pp. 233--240 (2006)

\bibitem{deng2018image}
Deng, W., Zheng, L., Ye, Q., Kang, G., Yang, Y., Jiao, J.: Image-image domain
  adaptation with preserved self-similarity and domain-dissimilarity for person
  re-identification. In: CVPR (2018)

\bibitem{ester1996density}
Ester, M., Kriegel, H.P., Sander, J., Xu, X., et~al.: A density-based algorithm
  for discovering clusters in large spatial databases with noise. In: Kdd.
  vol.~96, pp. 226--231 (1996)

\bibitem{fan2018unsupervised}
Fan, H., Zheng, L., Yan, C., Yang, Y.: Unsupervised person re-identification:
  Clustering and fine-tuning. ACM Transactions on Multimedia Computing,
  Communications, and Applications (TOMM)  (2018)

\bibitem{fawcett2006introduction}
Fawcett, T.: An introduction to roc analysis. Pattern recognition letters
  \textbf{27}(8),  861--874 (2006)

\bibitem{Fu2019SelfsimilarityGA}
Fu, Y., Wei, Y., Wang, G., Zhou, X., Shi, H., Huang, T.S.: Self-similarity
  grouping: A simple unsupervised cross domain adaptation approach for person
  re-identification. ICCV  (2019)

\bibitem{fu2019horizontal}
Fu, Y., Wei, Y., Zhou, Y., et~al.: Horizontal pyramid matching for person
  re-identification. In: AAAI (2019)

\bibitem{ge2018fd}
Ge, Y., Li, Z., Zhao, H., et~al.: Fd-gan: Pose-guided feature distilling gan
  for robust person re-identification. In: NeurIPS (2018)

\bibitem{grafarend2012linear}
Grafarend, E., Awange, J.: Linear and Nonlinear Models: Fixed Effects, Random
  Effects, and Total Least Squares. Springer (2012)

\bibitem{hermans2017defense}
Hermans, A., Beyer, L., Leibe, B.: In defense of the triplet loss for person
  re-identification. arXiv preprint arXiv:1703.07737  (2017)

\bibitem{hirzer2012relaxed}
Hirzer, M., Roth, P.M., K{\"o}stinger, M., Bischof, H.: Relaxed pairwise
  learned metric for person re-identification. In: ECCV. pp. 780--793. Springer
  (2012)

\bibitem{jia2020similarity}
Jia, M., Zhai, Y., Lu, S., Ma, S., Zhang, J.: A similarity inference metric for
  rgb-infrared cross-modality person re-identification. In: IJCAI (2020)

\bibitem{jin2020uncertainty}
Jin, X., Lan, C., Zeng, W., Chen, Z.: Uncertainty-aware multi-shot knowledge
  distillation for image-based object re-identification. In: AAAI (2020)

\bibitem{jin2020snr}
Jin, X., Lan, C., Zeng, W., Chen, Z., Zhang, L.: Style normalization and
  restitution for generalizable person re-identification. In: CVPR (2020)

\bibitem{jin2020semantics}
Jin, X., Lan, C., Zeng, W., Wei, G., Chen, Z.: Semantics-aligned representation
  learning for person re-identification. In: AAAI. pp. 11173--11180 (2020)

\bibitem{kingma2014adam}
Kingma, D.P., Ba, J.: Adam: A method for stochastic optimization. In: ICLR
  (2014)

\bibitem{kumar2016learning}
Kumar, B., Carneiro, G., Reid, I., et~al.: Learning local image descriptors
  with deep siamese and triplet convolutional networks by minimising global
  loss functions. In: CVPR. pp. 5385--5394 (2016)

\bibitem{li2018unsupervised}
Li, M., Zhu, X., Gong, S.: Unsupervised person re-identification by deep
  learning tracklet association. In: ECCV (2018)

\bibitem{li2014deepreid}
Li, W., Zhao, R., Tian, L., et~al.: Deepreid: Deep filter pairing neural
  network for person re-identification. In: CVPR (2014)

\bibitem{li2018adaptation}
Li, Y.J., Yang, F.E., Liu, Y.C., Yeh, Y.Y., Du, X., Frank~Wang, Y.C.:
  Adaptation and re-identification network: An unsupervised deep transfer
  learning approach to person re-identification. In: CVPR workshops (2018)

\bibitem{lin2018multi}
Lin, S., Li, H., Li, C.T., Kot, A.C.: Multi-task mid-level feature alignment
  network for unsupervised cross-dataset person re-identification. BMVC  (2018)

\bibitem{liu2019adaptive}
Liu, J., Zha, Z.J., Chen, D., Hong, R., Wang, M.: Adaptive transfer network for
  cross-domain person re-identification. In: CVPR (2019)

\bibitem{luo2019bag}
Luo, H., Gu, Y., Liao, X., Lai, S., Jiang, W.: Bag of tricks and a strong
  baseline for deep person re-identification. In: CVPR workshops (2019)

\bibitem{qi2019novel}
Qi, L., Wang, L., Huo, J., Zhou, L., Shi, Y., Gao, Y.: A novel unsupervised
  camera-aware domain adaptation framework for person re-identification. ICCV
  (2019)

\bibitem{qian2018pose}
Qian, X., Fu, Y., Wang, W., et~al.: Pose-normalized image generation for person
  re-identification. In: ECCV (2018)

\bibitem{ridgeway2018learning}
Ridgeway, K., Mozer, M.C.: Learning deep disentangled embeddings with the
  f-statistic loss. In: NeurIPS. pp. 185--194 (2018)

\bibitem{ristani2018features}
Ristani, E., Tomasi, C.: Features for multi-target multi-camera tracking and
  re-identification. In: CVPR. pp. 6036--6046 (2018)

\bibitem{song2020unsupervised}
Song, L., Wang, C., Zhang, L., Du, B., Zhang, Q., Huang, C., Wang, X.:
  Unsupervised domain adaptive re-identification: Theory and practice. Pattern
  Recognition p. 107173 (2020)

\bibitem{su2017pose}
Su, C., Li, J., Zhang, S., et~al.: Pose-driven deep convolutional model for
  person re-identification. In: ICCV (2017)

\bibitem{sun2018beyond}
Sun, Y., Zheng, L., Yang, Y., Tian, Q., Wang, S.: Beyond part models: Person
  retrieval with refined part pooling (and a strong convolutional baseline).
  In: ECCV. pp. 480--496 (2018)

\bibitem{szegedy2016rethinking}
Szegedy, C., Vanhoucke, V., Ioffe, S., Shlens, J., Wojna, Z.: Rethinking the
  inception architecture for computer vision. In: CVPR (2016)

\bibitem{tang2019unsupervised}
Tang, H., Zhao, Y., Lu, H.: Unsupervised person re-identification with
  iterative self-supervised domain adaptation. In: CVPR workshops (2019)

\bibitem{ustinova2016learning}
Ustinova, E., Lempitsky, V.: Learning deep embeddings with histogram loss. In:
  NeurIPS. pp. 4170--4178 (2016)

\bibitem{wang2018transferable}
Wang, J., Zhu, X., Gong, S., Li, W.: Transferable joint attribute-identity deep
  learning for unsupervised person re-identification. In: CVPR (2018)

\bibitem{wang2018resource}
Wang, Y., Wang, L., You, Y., Zou, X., Chen, V., Li, S., Huang, G., Hariharan,
  B., Weinberger, K.Q.: Resource aware person re-identification across multiple
  resolutions. In: CVPR (2018)

\bibitem{wei2018person}
Wei, L., Zhang, S., Gao, W., Tian, Q.: Person transfer {GAN} to bridge domain
  gap for person re-identification. In: CVPR (2018)

\bibitem{wojke2018deep}
Wojke, N., Bewley, A.: Deep cosine metric learning for person
  re-identification. In: WACV. pp. 748--756 (2018)

\bibitem{yang2019asymmetric}
Yang, F., Li, K., Zhong, Z., Luo, Z., Sun, X., Cheng, H., Guo, X., Huang, F.,
  Ji, R., Li, S.: Asymmetric co-teaching for unsupervised cross domain person
  re-identification. AAAI  (2020)

\bibitem{yang2019patch}
Yang, Q., Yu, H.X., Wu, A., Zheng, W.S.: Patch-based discriminative feature
  learning for unsupervised person re-identification. In: CVPR (2019)

\bibitem{ye2020deep}
Ye, M., Shen, J., Lin, G., Xiang, T., Shao, L., Hoi, S.C.: Deep learning for
  person re-identification: A survey and outlook. arXiv preprint
  arXiv:2001.04193  (2020)

\bibitem{yu2017cross}
Yu, H.X., Wu, A., Zheng, W.S.: Cross-view asymmetric metric learning for
  unsupervised person re-identification. In: ICCV (2017)

\bibitem{yu2018unsupervised}
Yu, H.X., Wu, A., Zheng, W.S.: Unsupervised person re-identification by deep
  asymmetric metric embedding. IEEE TPAMI  (2018)

\bibitem{yu2019unsupervised}
Yu, H.X., Zheng, W.S., Wu, A., Guo, X., Gong, S., Lai, J.H.: Unsupervised
  person re-identification by soft multilabel learning. In: CVPR (2019)

\bibitem{zhai2020ad}
Zhai, Y., Lu, S., Ye, Q., Shan, X., Chen, J., Ji, R., Tian, Y.: Ad-cluster:
  Augmented discriminative clustering for domain adaptive person
  re-identification. In: CVPR. pp. 9021--9030 (2020)

\bibitem{zhai2020multiple}
Zhai, Y., Ye, Q., Lu, S., Jia, M., Ji, R., Tian, Y.: Multiple expert
  brainstorming for domain adaptive person re-identification. In: ECCV (2020)

\bibitem{zhang2019self}
Zhang, X., Cao, J., Shen, C., You, M.: Self-training with progressive
  augmentation for unsupervised cross-domain person re-identification. In: ICCV
  (2019)

\bibitem{zhang2019DSA}
Zhang, Z., Lan, C., Zeng, W., et~al.: Densely semantically aligned person
  re-identification. In: CVPR (2019)

\bibitem{zhao2017spindle}
Zhao, H., Tian, M., Sun, S., et~al.: Spindle net: Person re-identification with
  human body region guided feature decomposition and fusion. In: CVPR (2017)

\bibitem{zheng2015scalable}
Zheng, L., Shen, L., et~al.: Scalable person re-identification: A benchmark.
  In: ICCV (2015)

\bibitem{zheng2016person}
Zheng, L., Yang, Y., Hauptmann, A.G.: Person re-identification: Past, present
  and future. arXiv preprint arXiv:1610.02984  (2016)

\bibitem{zheng2017person}
Zheng, L., Zhang, H., Sun, S., Chandraker, M., Yang, Y., Tian, Q.: Person
  re-identification in the wild. In: CVPR. pp. 1367--1376 (2017)

\bibitem{zheng2017unlabeled}
Zheng, Z., Zheng, L., Yang, Y.: Unlabeled samples generated by gan improve the
  person re-identification baseline in vitro. In: ICCV (2017)

\bibitem{zhong2017re}
Zhong, Z., Zheng, L., Cao, D., Li, S.: Re-ranking person re-identification with
  k-reciprocal encoding. In: CVPR (2017)

\bibitem{zhong2018generalizing}
Zhong, Z., Zheng, L., Li, S., Yang, Y.: Generalizing a person retrieval model
  hetero-and homogeneously. In: ECCV (2018)

\bibitem{zhong2019invariance}
Zhong, Z., Zheng, L., Luo, Z., Li, S., Yang, Y.: Invariance matters: Exemplar
  memory for domain adaptive person re-identification. In: CVPR. pp. 598--607
  (2019)

\bibitem{zhou2019omni}
Zhou, K., Yang, Y., Cavallaro, A., et~al.: Omni-scale feature learning for
  person re-identification. ICCV  (2019)

\bibitem{zhou2016deep}
Zhou, S., Wang, J., Hou, Q., Gong, Y.: Deep ranking model for person
  re-identification with pairwise similarity comparison. In: Pacific Rim
  Conference on Multimedia. pp. 84--94. Springer (2016)

\bibitem{zhu2017unpaired}
Zhu, J.Y., Park, T., Isola, P., et~al.: Unpaired image-to-image translation
  using cycle-consistent adversarial networks. In: ICCV (2017)

\end{thebibliography}
\end{document}